\documentclass[10pt,twocolumn,letterpaper]{article}

\usepackage{cvpr}              

\usepackage{multirow}
\usepackage{pifont}
\usepackage{colortbl}
\usepackage{xcolor}
\usepackage{booktabs}  
\usepackage{algorithm}
\usepackage{algpseudocode}
\usepackage[most]{tcolorbox}
\usepackage{bbm}

\definecolor{cvprblue}{rgb}{0.21,0.49,0.74}
\usepackage[pagebackref,breaklinks,colorlinks,allcolors=cvprblue]{hyperref}
\usepackage{balance}
\usepackage{multicol}

\def\paperID{} 
\def\confName{CVPR}
\def\confYear{2026}

\title{OSPO: Object-Centric Self-Improving Preference Optimization for Text-to-Image Generation}

\author{Yoonjin Oh, Yongjin Kim, Hyomin Kim, Donghwan Chi, Sungwoong Kim\thanks{Corresponding author}\\
Korea University\\
{\tt\small \{dhdbsrlw, rla020, khmiee, zheedong, swkim01\}@korea.ac.kr}}

\begin{document}
\maketitle
\begin{abstract}
Recent advances in Multimodal Large Language Models (MLLMs) have enabled unified multimodal understanding and generation. However, they still struggle with fine-grained text–image alignment, often failing to faithfully depict objects with correct attributes such as color, shape, and spatial relationships. To mitigate this issue, previous studies have explored preference optimization methods such as DPO and GRPO, but these approaches incur substantial computational cost, both in constructing preference data and in performing optimization. This has motivated self-improving preference optimization approaches, in which the MLLM autonomously generates its own training data, self-estimates preference feedback, and self-optimizes using the resulting self-constructed preference pairs. However, existing self-improving methods still overlook fine-grained, object-level semantics, allowing object hallucination to persist. To tackle this problem, we propose $\underline{\text{O}}$bject-centric $\underline{\text{S}}$elf-improving $\underline{\text{P}}$reference $\underline{\text{O}}$ptimization (OSPO), a self-improving framework designed to enhance object-level text–image alignment. OSPO explicitly constructs object-centric preference data without relying on any external data or models. We also introduce a new approach that leverages attention-based object masks together with an object-weighted SimPO loss to enhance object-specific fidelity. Extensive experiments on three compositional image generation benchmarks demonstrate that OSPO significantly improves fine-grained alignment and reduces object hallucination, outperforming prior self-improving methods and even specialized diffusion-based text-to-image models. The project page is available at \hyperlink{https://ku-agi.github.io/OSPO/}{https://ku-agi.github.io/OSPO/}.
\end{abstract}
    
\section{Introduction}
\label{sec:intro}

Recent advancements in Unified Multimodal Large Language Models (Unified MLLMs) \citep{team2024chameleon, sun2023emu1, sun2024emu2, wang2024emu3, ge2023seedllama, wu2024vilau, xie2024show1, xie2025show2} have achieved remarkable performance, demonstrating the ability to perform diverse vision tasks ranging from image understanding to generation within a single model. Despite these successes, a persistent challenge remains in achieving fine-grained alignment between text and image. In text-to-image (T2I) generation, MLLMs often fail to faithfully render the details specified in the input prompt, including spatial relationships and precise visual attributes of objects. A critical and related failure mode is object hallucination, which encompasses not only the generation of non-existent objects but also the omission or distortion of objects described in the prompt.

Early work has addressed fine-grained alignment and hallucination using feedback-based post-training methods such as DPO \citep{rafailov2023dpo, wang2024mdpo}, PPO \citep{schulman2017proximal} and GRPO \citep{shao2024deepseekmath}. Despite their effectiveness, these approaches suffer from two key limitations: high cost and off-policy optimization. DPO requires large amounts of preference pairs curated by humans or stronger external models; collecting such data for image generation is inherently far more complex and costly than for text generation. PPO and GRPO do not require pre-collected preference pairs, yet involve multiple models during training incurring substantial computational overhead. Furthermore, all three methods introduce off-policy bias from external sources: DPO is trained on preference pairs whose distribution diverges from the model's own outputs, while PPO and GRPO optimize against reward models trained on externally sourced data, creating a distributional mismatch that leads to unstable training.

To overcome these constraints, recent works have introduced self-improving approaches \citep{qu2024silmm, mao2025unirl, Hong2025SUDERSU} that eliminate reliance on external data or models by generating training data or reward signals internally. This direction is especially promising with the advent of Unified MLLMs, whose dual abilities in image understanding and generation enable a self-improving loop in which the model evaluates and refines its own outputs. Such a loop is particularly viable because MLLMs' visual comprehension has matured to the point where the model can serve as a reliable judge of its own generations. However, despite these strengths, current self-improving methods still lack mechanisms that explicitly enforce fine-grained text–image alignment, leaving them insufficient for fully resolving object-level hallucination and detailed alignment failures.


In this paper, we propose \textbf{Object-centric Self-improving Preference Optimization (OSPO)}, a self-improving framework that autonomously constructs fine-grained, object-focused T2I preference pairs and performs preference optimization guided by predicted object masks. Here, \textit{object-centric} refers to treating each object independently as a distinct unit of alignment, rather than treating the image holistically. This framework enables the model to better capture fine-grained text–image correspondences at the individual object level. OSPO operates in five stages: prompt generation, prompt perturbation and densification to create prompt pairs sharing global context but differing in fine-grained details, candidate image generation with object mask, preference pair construction based on Self-VQA, and preference optimization. 
We validate the effectiveness of OSPO across multiple MLLMs on three established benchmarks for fine-grained T2I generation. Notably, OSPO substantially improves fine-grained alignment, outperforming existing self-improving methods and even surpassing diffusion-based models specifically designed solely for image generation.

To summarize, our main contributions are as follows: 
\begin{itemize}
    \item We present OSPO, a five-stage self-improving framework that mitigates object hallucination in T2I generation without relying on either external datasets or auxiliary models.
    \item We propose a pipeline that constructs high-quality, object-centric preference data with shared global semantics but fine-grained local differences, and incorporates object-aware supervision using attention-based object masks with an object-weighted SimPO loss.
    \item We demonstrate the empirical effectiveness of OSPO with substantial object-level alignment improvements across multiple T2I benchmarks, surpassing prior self-improving and specialized diffusion models.
\end{itemize}


\section{Related Works}
\subsection{Multimodal Large Language Models (MLLMs)}
\label{sec:bg_mllms}
Early Multimodal Large Language Models (MLLMs), particularly large vision-language models \citep{liu2023llava, ye2023mplug, dai2023instructblip, chen2024internvl}, primarily extended pretrained LLMs for visual understanding by integrating vision encoders. Recent advances, however, have moved toward unified MLLMs that can both understand and generate visual content. This unification enables a single model to handle diverse tasks, ranging from image understanding to generation, within a shared set of model parameters. Among them, Discrete MLLMs \citep{ge2023seedllama, wu2024vilau, sun2023emu1, sun2024emu2, wang2024emu3, wu2025janus, chen2025januspro} represent visual information as quantized tokens produced by vector quantization (VQ) \citep{van2017neural, esser2021taming} or learned tokenizers. These models employ vector quantization to discretize images into tokens, thereby allowing both vision and language to be modeled uniformly through next-token prediction. In this work, we focus on discrete unified MLLMs that generate images as sequences of discrete tokens.

\subsection{Self-Improving Frameworks for T2I}
\label{sec:bg_self_improving}
Recent works have proposed self-improving preference optimization frameworks for T2I generation in MLLMs \citep{qu2024silmm, Hong2025SUDERSU, mao2025unirl}. These frameworks commonly adopt a \textit{Best-of-N sampling} strategy, which produces multiple candidate images from the same prompt and selects the highest- and lowest-scoring ones to form preference pairs.
While SILMM \citep{qu2024silmm}, the first such framework, targets compositional T2I generation—a task that inherently requires fine-grained object-level alignment—this objective is only partially reflected in its design, limited to the preference assignment stage. Subsequent frameworks \citep{Hong2025SUDERSU, mao2025unirl} jointly train on T2I generation and image captioning but do not address fine-grained alignment as an explicit objective.
Overall, existing frameworks lack object-centric design across their generation, preference assignment, and optimization stages, limiting their ability to address object hallucination. In contrast, our framework explicitly integrates object-level components throughout all stages, enabling more precise and semantically faithful T2I generation.
\section{Framework}
\label{sec:framework}

\begin{figure}[t!]
    \centering
    \includegraphics[width=\linewidth]{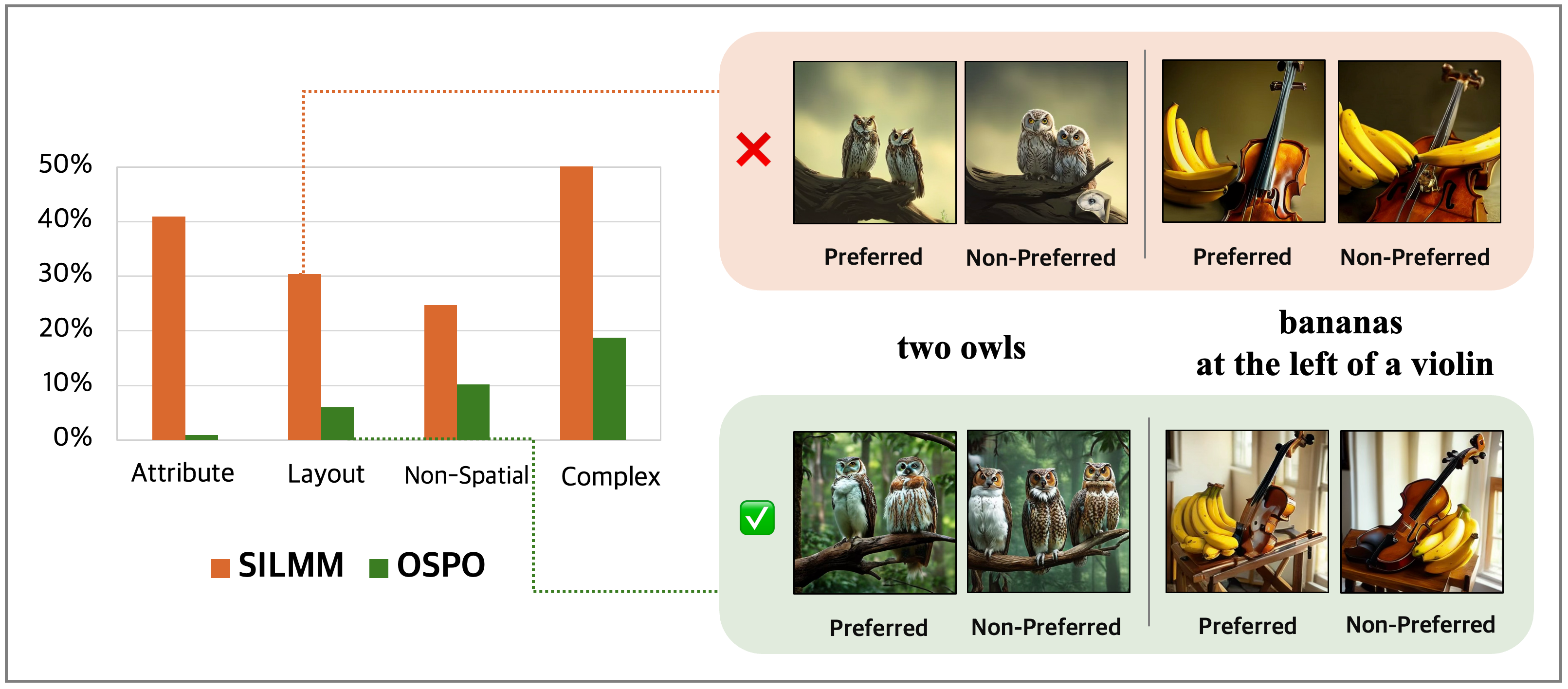}
    \caption{Proportion of Preference-Ambiguous Pairs across prompt categories in SILMM and OSPO (ours) datasets, along with representative examples. Lower is better.}
\label{fig:weak_preference}
\end{figure}

\begin{figure*}[t!]
    \centering
    \includegraphics[width=\linewidth]{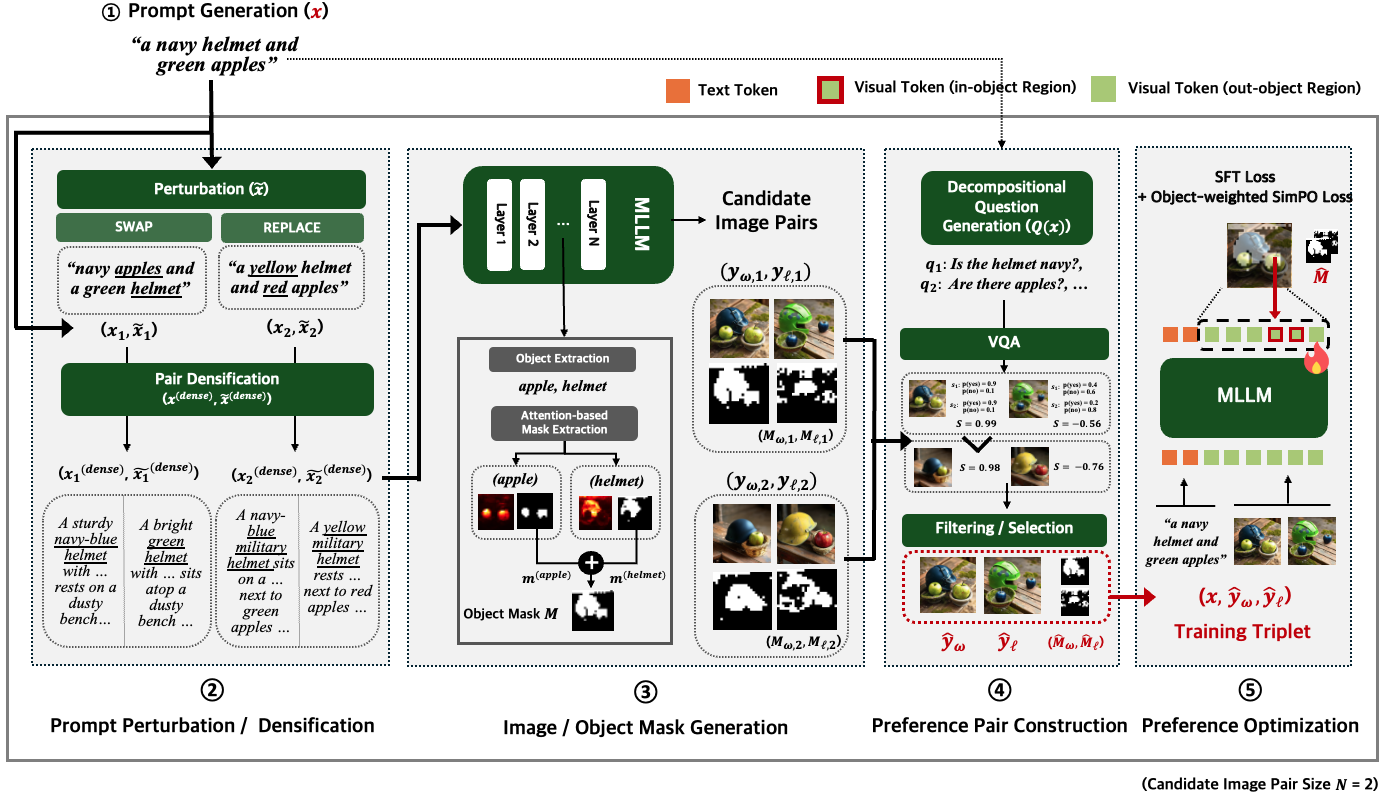}
    \caption{Overview of the OSPO framework: (Stage 1) The MLLM generates a set of base text prompts. (Stage 2)  For each base prompt, the model generates multiple perturbed variants using three strategies (Replace, Swap, and Drop) and each original–perturbed pair is jointly densified. Each pair of original and perturbed prompt is pairwise densified by the MLLM. (Stage 3) The MLLM generates candidate preferred and non-preferred images from each densified prompt pair. (Stage 4) The model constructs atomic decompositional VQA questions and evaluates each candidate image’s prompt fidelity using the Self-VQA alignment score S, filtering out noisy supervision and selecting a single final image pair. (Stage 5) The MLLM is fine-tuned using preference optimization with the proposed Object-weighted SimPO loss together with SFT loss.}
\label{fig:framework_overview}
\end{figure*}

\subsection{Preliminary}
\label{sec:preliminary}
We analyze the 16,000-sample self-generated preference dataset produced using Janus-Pro-7B \citep{chen2025januspro} following the workflow of SILMM \citep{qu2024silmm}. We identify a systematic failure mode of \textbf{Preference-Ambiguous Pair}, where both images exhibit indistinguishable levels of prompt fidelity, providing no meaningful preference signal for training. A particularly problematic subcase is \textbf{Preference-False Pair}, where both images are uniformly correct or uniformly incorrect across all semantic units. The forced preferred/non-preferred label assignment by \textit{Best-of-N sampling} actively introduces contradictory supervision. Figure~\ref{fig:weak_preference} reports the proportion of such preference-ambiguous pairs across prompt categories, revealing that this problem is prevalent across all categories. This indicates that \textit{Best-of-N sampling} is indiscriminate by nature and yields noisy preference assignments poorly aligned with the training objective. Formal definitions and the measurement protocol are provided in Appendix \ref{appendix:a-preliminary}.

\subsection{Overview}
We introduce \textbf{OSPO}, a five-stage self-improving preference optimization framework designed to mitigate object hallucination in T2I generation. The framework is fully self-contained, relying neither on external data nor pretrained reward models. By leveraging the dual capabilities of unified MLLMs, OSPO synthesizes text–image data from scratch without relying on \textit{Best-of-N sampling}. It then constructs object-centric preference pairs that deliver precise, fine-grained supervision, addressing limitations of prior self-improving frameworks and mitigating object-level hallucination. The five stages are as follows: (1) Prompt Generation (2) Prompt Perturbation and Densification (3)  Image and Object Mask Generation (4) VQA-based Preference Pair Construction (5) Preference Optimization. Figure \ref{fig:framework_overview} describes an overview of the OSPO framework.

\subsection{Prompt Generation}
\label{framework:stage1}
Inspired by T2I-Compbench++ \citep{huang2025t2icompbench}, we first construct the base set of training prompts, where each training prompt $x$ serves as the basis for subsequent data construction. The prompt set consists of four categories: Attribute, Layout, Non-spatial Relationship, and Complex Composition.
The Attribute category is decomposed into Color, Shape, and Texture, while Layout is decomposed into 2D and 3D-Spatial Relationships. The remaining categories are single-level categories. Non-spatial Relationship describes dynamic actions or states, while Complex Composition combines multiple categories to form richer and longer prompts. 
All prompts are generated by MLLMs via in-context learning. The prompt template is provided in Appendix \ref{appendix:h-prompt-templates}.

\subsection{Prompt Perturbation and Densification}
\label{framework:stage2}
To construct a preferred and non-preferred image pair $(y_w, y_\ell)$ for each text prompt $x$, we generate the two images independently rather than using the common \textit{Best-of-N sampling} strategy. We therefore generate preferred and non-preferred image variants using prompt pairs that share the same global semantics but differ slightly in object-level details, while leveraging the MLLM’s in-context learning capability. The templates are presented in Appendix \ref{appendix:h-prompt-templates}.




\subsubsection{Prompt Perturbation}
Inspired by SugarCrepe \citep{hsieh2023sugarcrepe} and WinoGround \citep{thrush2022winoground}, multiple perturbed prompts $\tilde{x}$ are generated for each text prompt $x$ using the following perturbation strategies: 
\begin{itemize}
    \item \textbf{Replace}: substitutes an object or attribute in the prompt with another that was not originally present, creating a novel compositional combination.
    
    \item \textbf{Swap}: exchanges the positions of objects or attributes within the prompt, inducing a different relational binding from the original.
    
    \item \textbf{Drop}: removes an object or attribute from the prompt, introducing semantic ambiguity.
\end{itemize}

 After generating $N$ perturbed prompts, each perturbed prompt is paired with the original text prompt to form $N$ distinct pairs: $(x, \tilde{x}_1)$, $(x, \tilde{x}_2)$, $\dots$, $(x, \tilde{x}_N)$. The original prompt $x$ is shared across all pairs, while each $\tilde{x}_{n}$, $n \in \{1, \dots, N\}$ represents a different perturbation. In our experiments, we set $N = 3$ by default.

\subsubsection{Prompt Densification}
Prior work \citep{yang2023idea2img, yang2025self, li2024promptist, wang2024promptcharm} shows that \textit{Prompt Densification}—adding contextual detail to prompts—improves faithfulness in T2I generation. We extend this idea by applying densification to each prompt pair $(x, \tilde{x})$ before image synthesis, ensuring that paired images share global semantics but differ only in fine-grained object details. Each prompt pair is densified jointly, producing two prompts with consistent background context but distinct object-level semantics. We analyze the impact of densification in Section~\ref{sec:analysis_dense}.

\subsection{Image and Object Mask Generation}
\label{framework:stage3}
In this stage, we generate candidate image pairs and their corresponding object masks, which are used in the preference optimization stage to introduce object-centric guidance during training. 
\subsubsection{Image Generation}
From each densified prompt pair $(x^{{(dense)}}, \tilde{x}^{(dense)})$, we generate one preferred and one non-preferred candidate image, denoted as $(y_{w}, y_{\ell})$. With $N$ prompt pairs, $N$ image pairs (a total of $2N$ images) are generated: $(y_{w,1}, y_{\ell,1})$, $(y_{w,2}, y_{\ell,2})$, $\dots$, $(y_{w,N}, y_{\ell,N})$. 

\subsubsection{Object Mask Generation}
During image generation, we extract a binary token-level mask, referred to as the \textit{Object Mask}, which indicates whether each visual token belongs to the object region of the image. This mask is obtained from attention weights, which makes the approach computationally efficient, as it does not require a separate segmentation model and leverages the model’s internal interactions \citep{kang2025your, binyamin2025make}. 

To obtain the object mask, we extract the attention distribution from a given object-representing text token over all visual tokens within the intermediate layers of the MLLM. Here, we exclude the top-$k$ earliest and top-$k$ latest layers to avoid unstable activations and over-smoothing. We then average them across heads and layers, and reshape the result into a 2D spatial attention map aligned with the image grid. This map is binarized into an object-specific foreground mask $m$ using OTSU's adaptive thresholding method \citep{otsu1975threshold}. We set $k=5$ by default. By repeating this process for all objects described in the original prompt $x$ and taking the union of the resulting masks, we obtain a single object mask $M$ for each generated image. Detailed formulations and algorithms are provided in Appendix \ref{appendix:b3-mask-extraction}.

\subsection{VQA-based Preference Pair Construction}
\label{framework:stage4}
Since preference optimization is highly sensitive to data quality, OSPO includes a pair filtering and selection stage to ensure the reliability of training pairs among the $N$ candidate image pairs. The overall process is guided by the Self-VQA results.

\subsubsection{Decompositional VQA}
\label{sec:decompositional-vqa}
 Decompositional Visual Question Answering (VQA) is a common method for fine-grained evaluation of text–image alignment \citep{huang2025t2icompbench, ghosh2023geneval, hu2024dpgbench, hu2023tifa}. It evaluates an image at the level of atomic semantic elements \citep{cho2023davidsonian}. This enables fine-grained evaluation across multiple semantic dimensions. In OSPO, every prompt $x$ in the base prompt set is decomposed by the MLLM into a set of binary Yes/No questions, $Q(x) = \{q_1, q_2, \ldots, q_K\}$, where $K$ denotes the number of atomic semantic elements. Each of the $N$ generated image pairs is evaluated using the corresponding question set, and the MLLM outputs Yes/No probabilities for every question. The alignment score $S(y)$ for an image $y$ is then computed as the average probability margin between the Yes and No responses: $s_k(y) = p(\textit{yes}\mid y,q_k)-p(\textit{no}\mid y,q_k), ~~
S(y)=\frac{1}{K}\sum_{k=1}^{K} s_k(y)$.

 


\subsubsection{Pair Filtering and Selection}
A candidate image pair $(y_w, y_{\ell})$ is filtered based on each image's per-question scores $s_1(\cdot),$ $\ldots,$ $s_K(\cdot)$ and their aggregated alignment score $S(\cdot)$ for both $y_w$ and $y_\ell$. Since the two images form a coupled pair, if either image is deemed invalid, the entire pair is discarded. In particular, the pair is discarded if the aggregated alignment score for the preferred image $y_w$ falls below the threshold: $S(y_w) < \tau$, where $\tau$ is a filtering threshold and we set $\tau=0.6$ by default. Similarly, the pair is discarded if the non-preferred image $y_{\ell}$ satisfies $s_k(y_{\ell}) > 0,\quad \forall k \in \{1, \dots, K\}$. By using both per-question scores $s_k$ and the aggregated score $S$, we remove preference-ambiguous and preference-false pairs to avoid noisy supervision. If no candidate pair survives this process, the entire text prompt and its generated images are excluded from training.

After filtering, the final training image pair $(\hat{y}_w, \hat{y}_\ell)$ is selected as the pair whose preferred image attains the highest alignment score $S$. Finally, we construct \textbf{preference training triplet: $(x, \hat{y}_w, \hat{y}_\ell)$.}





 
\subsection{Preference Optimization}
\label{framework:stage5}
Using the set of constructed preference triplets $\mathcal{D}$, we optimize the model using a combination of the Object-weighted SimPO loss and the Supervised Fine-Tuning (SFT) loss. Each loss term captures a complementary aspect of the preference learning objective.

\paragraph{Object-weighted SimPO Loss.}
Due to the nature of visual tokens, the standard SimPO objective \citep{meng2024simpo}, which aggregates reward over all tokens, can dilute the training signal by including many tokens that are irrelevant to the target object. Since our goal is fine-grained T2I alignment, we emphasize object-relevant visual tokens by applying spatial weights to the token-level reward in the standard SimPO.
The Object-weighted SimPO loss is defined as:
\begin{multline}
\mathcal{L}_{\text{Obj-SimPO}} = -\mathbb{E}_{(x,y_w,y_\ell)\sim\mathcal{D}}\Bigg[\log\sigma\bigg( \\
\frac{\beta}{|y_w|}
\sum_{t=1}^{|y_w|} {(w_w)}_t\log\pi_\theta\!\left((y_w)_t\mid x, {(y_w)}_{<t}\right) \\
-\,\frac{\beta}{|y_\ell|}
\sum_{t=1}^{|y_\ell|} {(w_\ell)}_t\log\pi_\theta\!\left((y_\ell)_t\mid x, {(y_\ell)}_{<t}\right)
- \gamma
\bigg)\Bigg],
\label{eq:simpo_object}
\end{multline}

where $\pi_\theta$ denotes the MLLM. $\sigma$, $\beta$, $\gamma$, and $w_t$ denote the sigmoid function, reward scaling factor, target reward margin, and token-level weighting factor computed by the object mask $m_t$, respectively. We define $w_t = 1 + \alpha\, m_t, \quad m_t \in [0,1]$, where $t$ denotes token position and $\alpha$ controls the emphasis on object-relevant visual tokens. This weighting scheme strengthens gradients on object-relevant tokens, producing sharper and more targeted preference signals. We set $\alpha=1$, $\beta=5$, $\gamma=2.5$ by default.  

%
 



\paragraph{SFT Loss.}
Visual token sequences encode spatial and structural patterns, where global coherence matters far more than the correctness of any single token. Unlike text generation, a token-level reward alone cannot effectively enforce object shape, geometry, or layout. Thus, leveraging only the PO objective may fail to provide sufficiently strong structural guidance. To complement this, we apply an SFT loss using preferred image as an anchor, ensuring consistent supervision over the full visual token sequence and reinforcing globally coherent image generation. The SFT loss is defined as:
\begin{align}
\mathcal{L}_{\text{SFT}}
&=
-\,\mathbb{E}_{(x,\,y_w) \sim \mathcal{D}}
\Bigg[
\frac{1}{|y_w|}
\sum_{t=1}^{|y_w|}
\log \pi_\theta\!\big((y_{w})_t \mid x,\, (y_{w})_{<t} \big)
\Bigg].
\end{align}

\paragraph{OSPO Loss.} The final training objective is a weighted combination of Object-weighted SimPO loss and SFT loss: $\mathcal{L}_{\text{OSPO}} = 
\mathcal{L}_\text{Obj-SimPO} 
 + \lambda \, \mathcal{L}_{\text{SFT}}$. Here, $\lambda$ is a loss weight for SFT loss, which we set $\lambda = 2$ by default.

\section{Experiment}
\label{sec:experiment}

\renewcommand{\arraystretch}{0.9}  
\begin{table*}[t!]
\centering
\fontsize{9}{11}\selectfont
\setlength{\tabcolsep}{2.3mm}
\caption{Comparison on T2I-CompBench++ \citep{huang2025t2icompbench}. 
↑ indicates higher is better, with bold highlighting the best score and underline indicating the second best among the unified MLLMs.}
{
\begin{tabular}{l c ccc c ccc c c}
\toprule
\multicolumn{1}{c}{\multirow{3}{*}{\textbf{Model}}} &
\multirow{3}{*}{\textbf{Size}} &
\multicolumn{3}{c}{\textbf{Attribute $\uparrow$}} &
\multicolumn{1}{c}{} &
\multicolumn{3}{c}{\textbf{Layout $\uparrow$}} &
\textbf{Non-} &
\textbf{Complex $\uparrow$} \\
\cmidrule(lr){3-5} \cmidrule(lr){7-9}
\multicolumn{1}{c}{} & & 
\textbf{Color} & \textbf{Shape} & \textbf{Texture} &
&
\textbf{Spatial-2D} & \textbf{Spatial-3D} & \textbf{Numeracy} &
\textbf{Spatial $\uparrow$} &
\textbf{} \\
\midrule
\multicolumn{11}{c}{\textbf{Diffusion Models}} \\
\midrule
SD-XL      & -- & 0.6369 & 0.5408 & 0.5637 & & 0.2032 & --     & --     & 0.3110 & 0.4091 \\
DALL-E 3   & -- & 0.7785 & 0.6205 & 0.7036 & & 0.2865 & --     & --     & 0.3003 & 0.3773 \\
FLUX.1     & -- & 0.7407 & 0.5718 & 0.6922 & & 0.2863 & --     & --     & 0.3127 & 0.3703 \\
\midrule
\multicolumn{11}{c}{\textbf{Unified MLLMs}} \\
\midrule
Show-o         & 7B & 0.5600 & 0.4100 & 0.4600 & & 0.2000 & --     & --     & 0.3000 & 0.2900 \\
Unitok-MLLM    & 7B & 0.7745 & 0.5195 & 0.6431 & & 0.2668 & \textbf{0.4088} & \underline{0.5675} & 0.3131 & 0.3617 \\
\midrule
Janus-Pro      & 1B & 0.3542 & 0.2291 & 0.2843 & & 0.0756 & 0.2367 & 0.2714 & 0.2809 & 0.2693 \\
+ SILMM        & 1B & 0.6097 & 0.3164 & 0.4474 & & 0.1539 & 0.2632 & 0.2826 & 0.3122 & 0.3679 \\
+ SUDER        & 1B & 0.7765 & 0.5106 & 0.6767 & & 0.2464 & -- & -- & 0.3130 & 0.3657 \\
\textbf{+ OSPO (ours)} & 1B &
\underline{0.8349} & 0.5763 & 0.7274 &
& \underline{0.3077} & \underline{0.4029} & 0.5653 &
\underline{0.3158} & \underline{0.3935} \\
\midrule
Janus-Pro      & 7B & 0.5215 & 0.3272 & 0.4050 & & 0.1654 & 0.2679 & 0.4431 & 0.3109 & 0.3868 \\
+ SILMM        & 7B & 0.7394 & 0.4325 & 0.5796 & & 0.2105 & 0.3572 & 0.5073 & 0.3113 & 0.3725 \\
+ SUDER        & 7B & 0.7824 & \underline{0.5786} & \underline{0.7292} & & 0.2524 & -- & -- & 0.3141 & 0.3858 \\
\textbf{+ OSPO (ours)} & 7B &
\textbf{0.8567} & \textbf{0.6386} & \textbf{0.7727} &
& \textbf{0.3562} & 0.3926 & \textbf{0.5908} &
\textbf{0.3161} & \textbf{0.4147} \\
\bottomrule
\end{tabular}
}
\label{tab:t2icompbench++}
\end{table*}

\subsection{Experiment Setting}
We conducted experiments using the Janus-Pro-1B and Janus-Pro-7B models \citep{chen2025januspro} as our primary backbones. For training, we used a dataset of 20,000 samples, obtained after filtering, spanning four categories: Attribute, Layout, Non-spatial Relationship, and Complex Composition.


\subsubsection{Benchmarks}
We evaluated T2I generation capabilities on the three standard benchmarks: T2I-CompBench++ \citep{huang2025t2icompbench}, DPGBench \citep{hu2024dpgbench}, and GenEval \citep{ghosh2023geneval}. Each evaluation was conducted following the standard configurations and protocols. Detailed descriptions of the evaluation methodology for each benchmark are provided in  Appendix \ref{appendix:d-evaluation-details}. 

\begin{itemize} 
    \item \textbf{T2I-CompBench++}: Evaluates compositional image generation ability using 2,400 prompts across four categories—Attribute, Layout, Non-spatial Relationship, and Complex Composition. 
    \item \textbf{DPGBench}: Evaluates image generation ability using 1,065 long, semantically rich prompts, which have an average token length of 83.91. It focuses on the semantic alignment of complex structures within extended narratives. 
    \item \textbf{GenEval}: Evaluates compositional image generation ability using 550 prompts in six categories using an evaluation protocol similar to T2I-CompBench++.
\end{itemize}

\subsubsection{Baselines} 
To evaluate the effectiveness of the OSPO framework, we compare OSPO with other self-improving preference optimization frameworks for T2I generation. Among self-improving frameworks, we include SILMM \citep{qu2024silmm} and SUDER \citep{Hong2025SUDERSU}, both of which construct preference pairs by generating images and assigning rewards without relying on external supervision. Note that SUDER has not released its model checkpoint, thus its results are omitted for certain categories or benchmarks.


\subsubsection{Implementation Details}
All experiments were conducted on eight NVIDIA A100 (80GB) GPUs. Additional implementation details, including training hyperparameters, are provided in Appendix \ref{appendix:c-implementation-details}.

\renewcommand{\arraystretch}{1.0}  
\begin{table*}[t!]
\centering
\fontsize{9}{11}\selectfont
\setlength{\tabcolsep}{0.95mm}
\caption{Comparison on DPGBench \citep{hu2024dpgbench} and GenEval \citep{ghosh2023geneval}. ↑ indicates higher is better, with bold indicating the best score and underline indicating the second best among  the unified MLLMs.}
{
\begin{tabular}{l c ccccc c ccccccc}
\hline
\multicolumn{1}{c}{\multirow{2}{*}{\textbf{Model}}} &
\multirow{2}{*}{\textbf{Size}} &
\multicolumn{6}{c}{\textbf{DPGBench $\uparrow$}} &
\multicolumn{7}{c}{\textbf{GenEval $\uparrow$}} \\
\cmidrule(r{0.5em}){3-8} \cmidrule(l{0.5em}){9-15}
\multicolumn{1}{c}{} & &
Attribute & Relation & Entity & Other & Global & \textbf{Overall} &
Single & Two & Count & Colors & Position &
\begin{tabular}[c]{@{}c@{}}Color\\ Attribution\end{tabular} &
\textbf{Overall} \\
\hline
\multicolumn{15}{c}{\textbf{Diffusion Models}} \\
\hline
SD-XL & - &
82.43 & 80.91 & 86.76 & 80.41 & 83.27 & 74.65 &
0.98 & 0.74 & 0.39 & 0.85 & 0.15 & 0.15 & 0.55 \\

DALL-E 3 & - &
89.61 & 88.39 & 90.58 & 89.83 & 90.97 & 83.50 &
0.96 & 0.87 & 0.47 & 0.83 & 0.43 & 0.43 & 0.67 \\

SD3-Medium & - &
- & - & - & - & - & - &
0.99 & 0.94 & 0.72 & 0.89 & 0.33 & 0.60 & 0.74 \\
\hline
\multicolumn{15}{c}{\textbf{Unified MLLMs}} \\
\hline
Show-o2 & 7B &
\underline{89.96} & 91.81 & \underline{91.78} & \textbf{91.64} & 89.00 & \textbf{86.14} &
\textbf{1.00} & 0.87 & 0.58 & \textbf{0.92} & 0.52 & 0.62 & 0.76 \\

Emu3 & 8B &
86.33 & 90.61 & 87.17 & \underline{89.75} & 87.54 & 81.60 &
0.98 & 0.69 & 0.33 & 0.78 & 0.15 & 0.16 & 0.52 \\

Unitok-MLLM & 7B &
88.37 & 91.39 & 88.13 & 87.54 & 83.98 & 81.87 &
\underline{0.99} & 0.75 & 0.39 & 0.80 & 0.21 & 0.41 & 0.59 \\
\hline

Janus-Pro & 1B &
88.17 & 88.98 & 88.63 & 88.30 & 87.58 & 82.63 &
\underline{0.99} & 0.82 & 0.48 & 0.89 & 0.62 & 0.57 & 0.73 \\

+ SILMM & 1B &
89.87 & \underline{91.91} & 88.67 & 82.60 & 78.22 & 83.56 &
\underline{0.99} & 0.86 & 0.48 & 0.89 & 0.70 & 0.62 & 0.75 \\

+ SUDER & 1B &
- & - & - & - & - & - &
\underline{0.99} & 0.87 & 0.54 & 0.90 &
0.74 & 0.62 & 0.78 \\

\textbf{+ OSPO (ours)} & 1B &
\textbf{90.42} & 87.24 & 89.09 & 87.08 & 89.21 & 84.46 &
\underline{0.99} & \underline{0.89} & 0.48 & \textbf{0.92} & 0.67 & 0.61 & 0.76 \\
\hline

Janus-Pro & 7B &
86.11 & \textbf{92.72} & 88.30 & 88.85 & 82.62 & 83.81 &
0.98 & 0.87 & 0.57 & 0.89 & 0.78 & 0.67 & 0.80 \\

+ SILMM & 7B &
89.47 & 91.26 & 90.33 & 88.37 & \textbf{91.46} & 84.56 &
0.98 & 0.87 & 0.58 & 0.90 & 0.77 & \underline{0.71} & 0.80 \\

+ SUDER & 7B &
- & - & - & - & - & - &
\underline{0.99} & \underline{0.89} & \textbf{0.70} & \textbf{0.92} &
\underline{0.82} & \underline{0.71} & \textbf{0.84} \\

\textbf{+ OSPO (ours)} & 7B &
89.03 & 89.45 & \textbf{92.14} & 84.22 & \underline{91.09} & \underline{85.61} &
\textbf{1.00} & \textbf{0.90} & \underline{0.63} & \underline{0.91} &
\textbf{0.83} & \textbf{0.72} & \underline{0.83} \\
\hline

\end{tabular}
}
\label{tab:dpg_geneval}
\end{table*}

\subsection{Main Results}
\label{sec:experiment_result}
As presented in Table \ref{tab:t2icompbench++}, the OSPO-trained model outperforms all self-improving baselines on T2I-CompBench++ for both the 1B and 7B model scales. Notably, substantial improvements are observed in the Attribute category, while the Non-spatial and Complex categories show relatively smaller improvements. This disparity arises from differences in the evaluation metrics across categories, particularly the CLIP-T score \citep{radford2021clip, hessel2021clipscore}, which is used for the Non-spatial and Complex categories, has limited sensitivity within its scoring range. In this context, comparisons with other methods and models show that our model performs comparably well in those categories. 

Our models of different sizes show consistent gains on other benchmarks, demonstrating the robustness of OSPO across scales. In Table \ref{tab:dpg_geneval}, our 7B model achieves the second-highest overall score among unified MLLMs on DPGBench, while attaining the highest overall score among Janus-Pro-based self-improving methods. Moreover, it achieves the second-best overall score on the GenEval benchmark with only a marginal performance gap. This gap mainly comes from the Count category, a known weakness of MLLMs. While SUDER outperforms ours in this category, this is largely attributable to its use of external COCO \citep{chen2015microsoft} image-text pair data, which provides counting supervision beyond what a purely self-improving framework can access. We also observe isolated cases where the 1B model outperforms the 7B model in specific categories; however, these inversions are category-specific, and the 7B model demonstrates more balanced gains across categories, resulting in superior overall performance. 
Overall, results in Table \ref{tab:dpg_geneval} demonstrate strong generalization ability of our model to out-of-distribution input prompts. Additionally, we evaluate general T2I generation performance beyond compositional ability using CLIP and LPIPS score in Appendix \ref{appendix:e3-clip-lpips}. Qualitative examples generated by the OSPO-trained model are shown in Figure \ref{fig:qualitative}, with additional samples provided in Appendix~\ref{appendix:g-qualitative}.



\begin{figure}[t!]
    \centering
    \includegraphics[width=\linewidth]{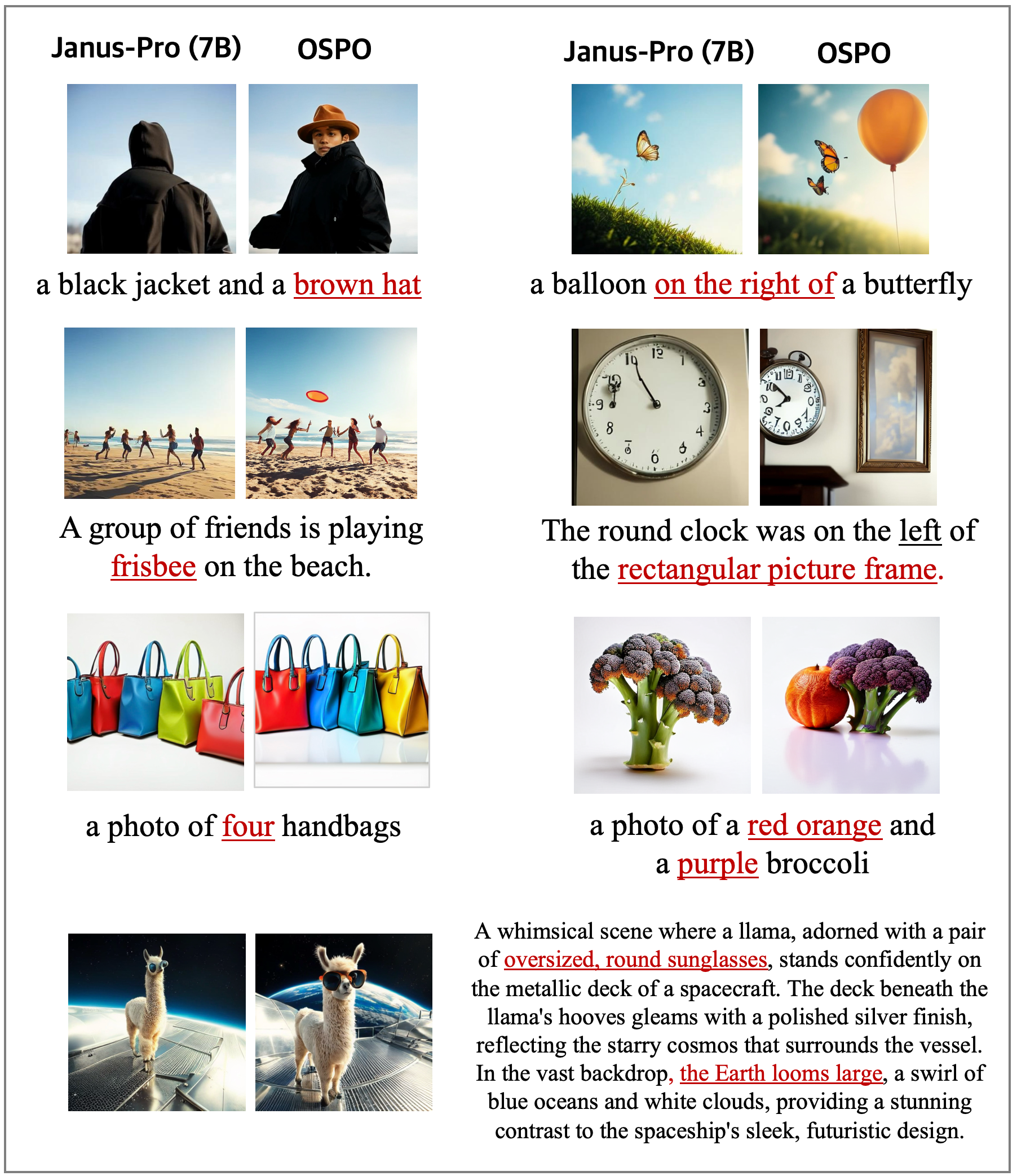}
    \caption{Qualitative examples from Janus-Pro-7B + OSPO on T2I-CompBench++, GenEval, and DPGBench prompts.}
\label{fig:qualitative}
\end{figure}



\section{Analysis}
\label{sec:analysis}
We conducted several experiments to analyze the contributions of individual components within the OSPO framework. All experiments were conducted using the Janus-Pro-7B model, focusing on three key aspects: (1) preference data (2) training loss and (3) computational cost. Appendix~\ref{appendix:e-analyses} provides additional analyses including iterative training, alternative 7B backbones, and data-generation biases.

\subsection{Object-centric Preference Data Construction}
OSPO is a self-improving framework that generates object-centric preference data that provides clear preference signals effective for training, as shown in Figure \ref{fig:weak_preference} representing a low proportion of preference-ambiguous pairs. It generates its own training data from scratch, enabling the data volume to scale with the available compute. Therefore, we first investigate two factors that influence its performance: the sample size, which refers to the total number of text prompts $x$ (see Section~\ref{framework:stage1}), and the candidate image pair size $N$ (see Sections~\ref{framework:stage2} and \ref{framework:stage3}), which refers to the number of image pairs produced per prompt. 


\begin{figure}[t!]
    \centering
    \includegraphics[width=\linewidth]{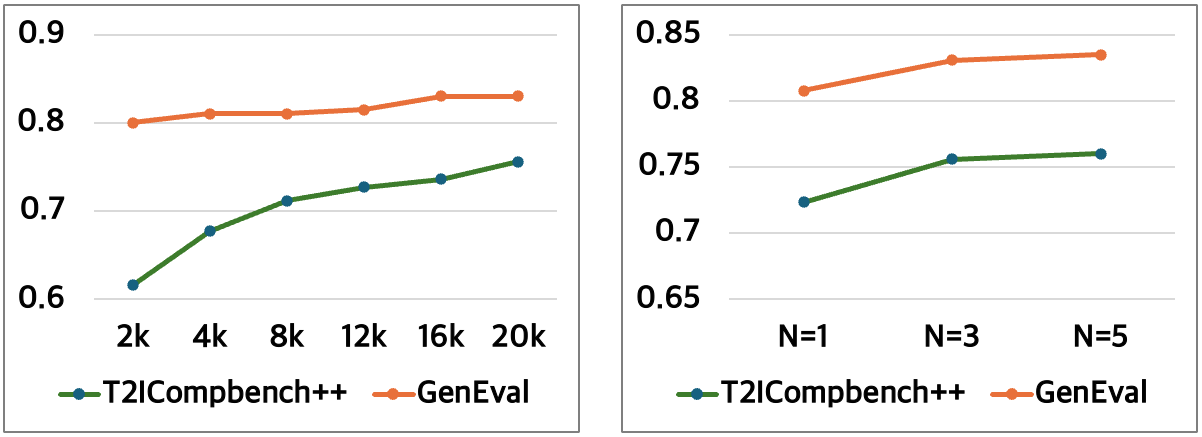}
    \caption{(Left) Effect of sample size on T2I-Compbench++ (attribute) and GenEval (overall) scores. (Right) Effect of candidate image pair size on T2I-Compbench++ (attribute) and GenEval (overall) scores. Higher is better for all benchmarks.}
\label{fig:effect_data_size}
\end{figure}

\subsubsection{Sample Size}
\label{sec:analysis_sample_size}
As shown in Figure~\ref{fig:effect_data_size}, model performance consistently improves as the total number of training samples increases. Notably, even with a minimal dataset, the model still achieves a substantial performance gain over the baseline. This highlights the data efficiency of OSPO framework.

\subsubsection{Candidate Image Pair Size}
\label{sec:analysis_pair_size}
We further analyze the effect of the candidate image pair size on overall performance. As shown in Figure~\ref{fig:effect_data_size}, sampling multiple candidate pairs per prompt leads to consistent performance improvements compared to using a single pair. Although the performance gain gradually stabilizes with larger pair sizes, the results suggest that a moderate number of candidates is sufficient to capture diverse and informative training signals.

\subsubsection{Image Sampling and Pair Filtering/Selection}
\label{sec:analysis_dense}


\paragraph{Effect of Prompt Densification on Image Fidelity.}
\label{sec:abl-wo-dense}
To verify that densified prompts with richer semantic descriptions really lead to more accurate text–image alignment and better model performance, we conducted a comparative study between two settings: generating images from the original prompt $x$ and from its densified version $x^{(dense)}$. As shown in Table~\ref{tab:densification_curation}, training the model with OSPO including the densification stage consistently yields higher performance on both benchmarks. Therefore, these results validate the effectiveness of the densification process. Also, the results of text-image alignment evaluation are provided in Appendix~\ref{appendix:e5-densification}.

\begin{table}[t!]
\centering
\fontsize{8.5}{10}\selectfont
\setlength{\tabcolsep}{0.9mm}
\caption{Effect of filtering and selection strategy in two different settings (with and without densification) construction stage on T2I-Compbench++ (attribute) and GenEval (overall) scores. Higher is better for all benchmarks.}
\begin{tabular}{lcccc}
\toprule
\textbf{Model} & \textbf{Filtering} & \textbf{Selection} & \textbf{T2I++}$\uparrow$ & \textbf{GenEval}$\uparrow$ \\
\midrule
\textbf{Janus-Pro-7B} &  &  & 0.418 & 0.796 \\
\midrule
\multirow{4}{*}{\textbf{OSPO w/ Densification}}
  & \ding{55} & \ding{55} & 0.716 & 0.813 \\
  & \ding{51} & \ding{55} & 0.725 & 0.810 \\
  & \ding{55} & \ding{51} & 0.724 & 0.805 \\
  & \ding{51} & \ding{51} & \textbf{0.756} & \textbf{0.831} \\
\midrule
\multirow{4}{*}{\textbf{OSPO w/o Densification}}
  & \ding{55} & \ding{55} & 0.618 & 0.816 \\
  & \ding{51} & \ding{55} & 0.549 & 0.814 \\
  & \ding{55} & \ding{51} & 0.627 & 0.814 \\
  & \ding{51} & \ding{51} & \textbf{0.641} & \textbf{0.823} \\
\bottomrule
\end{tabular}
\label{tab:densification_curation}
\end{table}

\paragraph{Effect of Filtering and Selection Strategy.}
We further analyze the interaction between the filtering–selection stage and the prompt densification stage. These two stages are closely interdependent: the effectiveness of the filtering and selection processes largely depends on the diversity and semantic richness of the generated image samples.
To capture this dependency, we investigate the combined impact of densification and filtering on benchmark performance, highlighting how the quality and distribution of generated samples directly influence the downstream preference pair selection process. Table \ref{tab:densification_curation} shows that the filtering and selection strategies consistently improve benchmark performance, regardless of whether prompt densification is applied. Notably, the effect becomes more pronounced when prompts are not densified and images are generated directly. This indicates that filtering and selection play a more critical role in mitigating noise and ensuring data quality when training on images with incomplete semantic fidelity. Additional experiments regarding the filtering threshold are provided in Appendix~\ref{appendix:e6-tau}.

\subsection{Training Loss Design for Object-centric PO}
\begin{table}[t!]
\centering
\fontsize{9}{11}\selectfont
\setlength{\tabcolsep}{1.2pt}
\caption{Ablation study on loss components with T2I-Compbench++ (attribute, layout) and GenEval (overall, position) scores. Higher is better for all benchmarks.}
\begin{tabular}{lcccccccc}
\toprule
\multicolumn{1}{c}{\multirow{2}{*}{\textbf{Model}}} &
\multicolumn{1}{c}{\multirow{2}{*}{$\mathcal{L}_{\text{SimPO}}$}} &
\multicolumn{1}{c}{\multirow{2}{*}{$\mathcal{L}_{\text{Obj-SimPO}}$}} &
\multicolumn{1}{c}{\multirow{2}{*}{$\mathcal{L}_{\text{SFT}}$}} &
\multicolumn{2}{c}{\textbf{T2I++}$\uparrow$} &
\multicolumn{2}{c}{\textbf{GenEval}$\uparrow$} \\
\cmidrule(lr){5-6} \cmidrule(lr){7-8}
& & & &
Attr. & Layout &
Over. & Position \\
\midrule
\textbf{Janus-Pro-7B} & & & & 0.418 & 0.292 & 0.796 & 0.570 \\
\midrule
 & \ding{51} & & & \textbf{0.779} & 0.416 & 0.785 & 0.778 \\
 & & \ding{51} & & 0.776 & 0.428 & 0.794 & 0.795 \\
 & \ding{51} & & \ding{51} & 0.755 & 0.437 & 0.827 & 0.795 \\
\textbf{+ OSPO (ours)} & & \ding{51} & \ding{51} & 0.756 & \textbf{0.447} & \textbf{0.831} & \textbf{0.828} \\
\bottomrule
\end{tabular}
\label{tab:loss_ablation}
\end{table}

As shown in Table \ref{tab:loss_ablation}, both loss components contribute to consistent performance gains across benchmarks. In particular, improvements are notable in spatial text–image alignment (the \textit{Layout} category in T2I-CompBench++ and the \textit{Position} category in GenEval). This suggests that the Object-weighted SimPO loss effectively focuses learning on object-relevant regions, while the SFT loss provides complementary structural supervision by acting as a stabilizing anchor during training. Together, these objectives enhance the model’s text-faithful and spatially coherent image synthesis.


\subsection{Computational Cost}
\label{sec:analysis_cost}
\begin{figure}[t!]
    \centering
    \includegraphics[width=\linewidth]{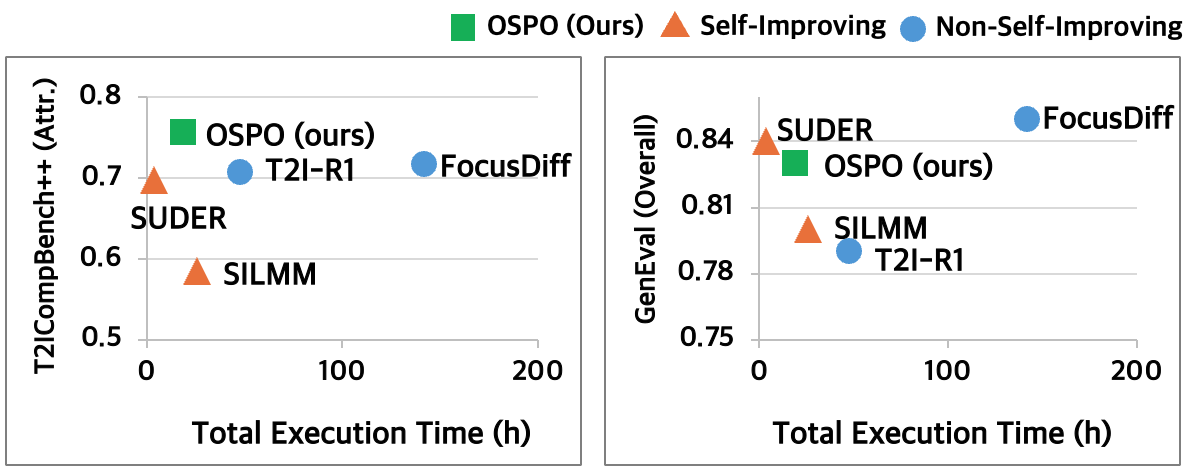}
    \caption{Performance–Efficiency trade-off across preference optimization frameworks for T2I generation. (Left) Comparison of T2I-CompBench++ (attribute) scores against execution cost. (Right) Comparison of GenEval (overall) scores against execution cost. Higher is better for all benchmarks.}
\label{fig:time_cost}
\end{figure}

Finally, we analyze the time cost of the OSPO framework compared with other preference optimization frameworks including both self-improving frameworks \citep{qu2024silmm, Hong2025SUDERSU} and non–self-improving frameworks \citep{jiang2025t2ir1, pan2025focusdiff}. The time cost was measured on a system equipped with eight NVIDIA A100 (80 GB) GPUs for one iteration of each framework. 
We first compare OSPO with SILMM \citep{qu2024silmm}. As shown in Figure~\ref{fig:time_cost},  OSPO achieves higher benchmark performance than SILMM \citep{qu2024silmm} while being more time-efficient, primarily because it generates smaller, targeted candidate sets by decoupling the source prompt.
Compared with T2I-R1 \citep{jiang2025t2ir1} and FocusDiff \citep{pan2025focusdiff}, which rely on GRPO \citep{shao2024deepseekmath} with multiple reward models or combine SFT and GRPO with image-editing datasets for fine-grained alignment, OSPO achieves comparable performance at substantially lower computational cost. A detailed breakdown of the wall-clock time for each stage of OSPO is provided in Appendix~\ref{appendix:b1-time}.



\section{Conclusion}
\label{sec:conclusion}

We propose an \textbf{\underline{O}}bject-centric \textbf{\underline{S}}elf-improving \textbf{\underline{P}}reference \textbf{\underline{O}}ptimization (\textbf{OSPO}) framework for T2I generation with MLLMs. OSPO leverages the unified multimodal capabilities of MLLMs to self-enhance fine-grained text–image alignment. By self-generating object-centric preference data that emphasize fine-grained semantic differences and applying object-weighted preference optimization, the model effectively self-improves its ability to capture detailed object-level semantics. Experiments on compositional image generation benchmarks show that OSPO greatly improves fine-grained alignment in T2I generation. 


\newpage
\section{Acknowledgements}
\label{sec:ack}

This work was supported by the National Research Foundation of Korea(NRF) grant funded by the Korea government(MSIT) (No. RS-2024-00353007, 20\%) and the Institute of Information \& Communications Technology Planning \& Evaluation(IITP) grant funded by the Korea government(MSIT) (IITP-2026-RS-2024-00436857, ITRC(Information Technology Research Center), 20\%), (No. RS-2019-II190079, Artificial Intelligence Graduate School Program(Korea University), 20\%), (IITP-2026-RS-2025-02304828, artificial intelligence star fellowship support program to nurture the best talents, 20\%), and (RS-2025-02653113, High-Performance Research AI Computing Infrastructure Support at the 2 PFLOPS Scale, 20\%).

\appendix 
\clearpage
\setcounter{page}{1}
\maketitlesupplementary

\section{Preliminary Details}
\label{appendix:a-preliminary}
This section provides additional explanations of the definitions of Preference-Ambiguous Pairs and Preference-False Pairs introduced in Section \ref{sec:preliminary} of the main paper. Along with that, we provide quantitative method used to measure their proportions in the dataset using per-question alignment score $s$ (defined in Section~\ref{sec:decompositional-vqa} of the main paper). 

Let define the binary indicator $\mathbb{I}_k(y) = \mathbbm{1}\{ s_k(y) > 0 \}$,
where $s_k(y)$ denotes per-question score of image $y$ on the $k$-th atomic question.


\begin{itemize}
    \item \textbf{Preference-Ambiguous Pair}: A pair $(y_w, y_\ell)$ is a \textit{preference-ambiguous} pair if the two images exhibit the same correctness pattern across all semantic units: $\forall k,\quad \mathbb{I}_k(y_w) = \mathbb{I}_k(y_\ell).$
    In this case, both images are either correct or incorrect on the same semantic units, providing no meaningful preference signal. Thus, the distinction between preferred and non-preferred becomes ambiguous.
    
    \item \textbf{Preference-False Pair}: A pair $(y_w, y_\ell)$ is a \textit{preference-false} pair if both images are either entirely correct or entirely incorrect across all semantic units: $(\forall k,\quad \mathbb{I}_k(y_w) = \mathbb{I}_k(y_\ell) = 0)$ or $(\forall k,\quad \mathbb{I}_k(y_w) = \mathbb{I}_k(y_\ell) = 1)$. This occurs when prompts are too simple (all correct) or too difficult (all wrong), causing \textit{Best-of-N sampling} to force a label assignment despite no meaningful quality difference, actively introducing contradictory supervision. Preference-false pairs form a strict subset of preference-ambiguous pairs.
    
\end{itemize}



\section{Framework Details}
\label{appendix:b-framework-details}

\subsection{Prompt Perturbation}
\label{appendix:b2-prompt-perturbation}
In the prompt perturbation stage, we apply different perturbation strategies depending on the prompt category. For Attribute and Layout prompts that contain a single object (e.g., “a green dress”, “two trees”), we restrict perturbations to the Replace strategy and generate $K$ variants by changing only the random seed. For Non-spatial prompts, we similarly limit perturbations to Replace and Drop, since other strategies are less meaningful for this category. For all other prompts, including multi-object Attribute/Layout prompts and Complex prompts, we randomly sample from all three perturbation strategies.

\begin{algorithm}[t]
\caption{Object-Centric Token Weighting}
\label{alg:object_token_weighting}
\begin{algorithmic}[1]

\State \textbf{Input:}
    Binary object mask $M$,
    Gaussian kernel $G$, 
    Emphasis strength $\alpha$
    
\State \textbf{Output:}
    Weight vector $\mathbf{w} \in \mathbb{R}^{T^2}$, 
    where $T^2$ denotes the number of visual tokens per image
\State $\tilde{M} \gets G * M$   \Comment{Gaussian smoothing}
\State $\hat{M} \gets \dfrac{\tilde{M} - \min(\tilde{M})}{\max(\tilde{M}) - \min(\tilde{M})}$ \Comment{Normalization}
\State $\mathbf{m} \gets \text{Flatten}(\hat{M})$
\State $\mathbf{w} \gets \mathbf{1} + \alpha \cdot \mathbf{m}$  
\State \Return $\mathbf{w}$

\end{algorithmic}
\end{algorithm}

\subsection{Object Mask Extraction}
\label{appendix:b3-mask-extraction}
\paragraph{Attention Map Extraction.} 
To obtain object masks, we begin by extracting all noun phrases from the input prompt using the spaCy linguistic parser, where each noun phrase is treated as an object candidate. For each object token (or token span), we probe the model’s self-attention mechanism during image generation.

Let $\mathcal{L}$ be the set of selected transformer layers—excluding the earliest and latest layers—from which attention maps are extracted.
For each layer $\ell \in \mathcal{L}$, we obtain the multi-head self-attention weights
$A_{\ell} \in \mathbb{R}^{|\mathcal{H}| \times (L+T^2) \times (L+T^2)}$,
where $L$ and $T^2$ denotes the number of text tokens and visual tokens, respectively, and $\mathcal{H}$ denotes the set of attention heads with $|\mathcal{H}|$ heads.

Since we are interested in attention from text tokens to visual tokens, we extract the query-to-visual slice for each head $h \in \mathcal{H}$. Let the visual-key index set be
$\mathcal{V} = \{L+1, \dots, L+T^2\}$. For each head h, the attention from any query token $i$ to the visual tokens is: $A_{\ell}(h)[i, \mathcal{V}] \in \mathbb{R}^{T^2}$. Let $S \subseteq \{1, \dots, L\}$ be the set of query token indices corresponding to a particular object span (possibly a single token or a multi-token phrase).
We compute the object-level attention for head $h$ at layer $\ell$ by averaging the attention vectors of all query tokens in $S$:

\begin{eqnarray}
A_{\ell,h}^{(S)}
=
\frac{1}{|S|}
\sum_{i \in S}
A_{\ell}(h)[i,\mathcal{V}]
\quad\in\mathbb{R}^{T^2}.
\end{eqnarray}
This yields a single attention distribution over all $T^2$ visual tokens for the object, per head and per layer.

\begin{table*}[t]
\centering
\caption{Comparison of training configurations for Janus-Pro-1B, Janus-Pro-7B, and UniTok-MLLM-8B.}
\fontsize{9}{11}\selectfont
\setlength{\tabcolsep}{6.8mm}
\begin{tabular}{@{}lccc@{}}
\toprule
\textbf{Configuration} & \textbf{Janus-Pro-1B} & \textbf{Janus-Pro-7B} & \textbf{UniTok-MLLM-8B} \\
\midrule
LoRA Rank & 64 & 64 & 32 \\
LoRA Alpha & 128 & 128 & 64 \\
Optimizer & AdamW & AdamW & AdamW \\
Optimizer Hyperparameters & $\beta_1$: 0.9, $\beta_2$: 0.95, $\epsilon$: 1e-8 & $\beta_1$: 0.9, $\beta_2$: 0.95, $\epsilon$: 1e-8 & $\beta_1$: 0.9, $\beta_2$: 0.95, $\epsilon$: 1e-8 \\
Gradient Clipping & 1.0 & 1.0 & 1.0 \\
Weight Decay & 0.0 & 0.0 & 0.0 \\
Learning Rate & 1e-5 & 1e-5 & 6e-6 \\
Learning Rate Scheduler & Cosine & Cosine & Constant \\
Total Training Steps & 500 & 800 & 300 \\
Total Batch Size & 128 & 128 & 128 \\
SimPO Hyperparameters & $\beta$: 5.0, $\gamma$: 2.5 & $\beta$: 5.0, $\gamma$: 2.5 & $\beta$: 3.0 , $\gamma$: 2.5 \\
SFT Loss Weight ($\lambda$) & 2.0 & 2.0 & 2.0 \\
Mixed Precision & bf16 & bf16 & bf16 \\
\bottomrule
\end{tabular}
\label{tab:training_settings_comparison}
\end{table*}


\paragraph{Attention Map Aggregation.}

Next, we average the object-conditioned attention weights across all heads $h \in \mathcal{H}$ and layers $\ell \in \mathcal{L}$ to obtain a single attention distribution over the $T^2$ visual tokens:
\begin{eqnarray}
\label{eq:attn_layer_head}
\bar{A}^{(S)}
=
\frac{1}{|\mathcal{L}|\,|\mathcal{H}|}
\sum_{\ell \in \mathcal{L}}
\sum_{h \in \mathcal{H}}
A_{\ell,h}^{(S)}
\;\in\mathbb{R}^{T^2}.
\end{eqnarray}


The attention distribution is then min–max normalized to $[0,1]$ and binarized using Otsu’s thresholding \citep{otsu1975threshold}. The aggregated attention map is binarized by applying Otsu's threshold $\tau_{\mathrm{otsu}}$:
\begin{eqnarray}
M^{(\mathcal{S})} =
\begin{cases}
1, & \bar{A}^{(\mathcal{S})} \ge \tau_{\mathrm{otsu}}, \\
0, & \text{otherwise}.
\end{cases}
\end{eqnarray}

\subsection{Object Mask Application}
\label{appendix:b3-mask-application}
During training, the binary Otsu mask $M$ is transformed into a continuous per-token mask $\mathbf{m} \in [0, 1]^{T^2}$ through Gaussian smoothing and min-max normalization, as described in Algorithm~\ref{alg:object_token_weighting}. The resulting weight vector $\mathbf{w}$ modulates the contribution of each visual token to the loss: object-relevant tokens receive larger weights while background tokens receive smaller weights, with a smooth transition between them. The weighting is defined solely over visual tokens — text tokens do not participate in the log-probability computation. In effect, the weighted token sequence amplifies gradients on object-relevant visual regions in both preferred and rejected outputs, guiding the preference optimization to focus on correcting object-specific errors.

\section{Implementation Details}
\label{appendix:c-implementation-details}

\subsection{Data Construction} 
\label{appendix:c1-text-hparam}

From stage 1 to stage 4, all text and image data are generated using a single MLLM. For text generation, stage 1 (Prompt Generation) adopts stochastic sampling with temperature=1.0 and top-p=0.9 rather than greedy decoding, and further encourages diversity by periodically rotating the in-context demonstration set throughout generation. Stage 2 (Prompt Perturbation and Densification) uses the same sampling configuration. 
For stage 3 (Image and Object Mask Generation), we set temperature=1.0 and cfg-weight=5.0. Lastly for stage 4 (VQA-based Preference Pair Construction), we use a single beam with temperature=1.2 and top-p=0.5 for question generation and answering.

\subsection{Preference Optimization}
\label{appendix:c2-train-hparam}

We provide detailed training configurations for OSPO in this section. We employ Low-Rank Adaptation (LoRA) \citep{hu2022lora} method for efficient optimization. Table~\ref{tab:training_settings_comparison} presents the training configurations for the two main experimental model sizes, Janus-Pro-1B and Janus-Pro-7B~\citep{chen2025januspro}, as well as the additional backbone model Unitok-MLLM-8B \citep{ma2025unitok}, which is discussed in Section \ref{appendix:e4-backbone}. 

\begin{table}[t]
\centering
\caption{Stage-wise wall-clock time breakdown of OSPO for a single iteration.}
\label{tab:time_breakdown}
\setlength{\tabcolsep}{0.8mm}
\begin{tabular}{lc}
\toprule
\textbf{Stage} & \textbf{Time (h)} \\
\midrule
(1) Prompt Generation                       & 0.07  \\
(2) Prompt Perturbation and Densification   & 5.06  \\
(3) Image and Object Mask Generation        & 6.80  \\
(4) VQA-based Preference Pair Construction  & 6.45  \\
(5) Preference Optimization                 & 1.83  \\
\midrule
\textbf{Total}                          & \textbf{20.21} \\
\bottomrule
\end{tabular}
\end{table}
\subsection{Stage-wise Time Cost Breakdown}
\label{appendix:b1-time}
Table~\ref{tab:time_breakdown} presents the wall-clock time for each stage of OSPO over a single iteration, measured on eight NVIDIA A100 (80GB) GPUs. In the default configuration, where 25,000 text prompts are generated in stage 1 and a pair size of 3 is employed in the subsequent stages, the total runtime amounts to 20.21 hours. 

\section{Evaluation Benchmark Details}
\label{appendix:d-evaluation-details}

This section describes the evaluation benchmarks in detail, focusing on how generated images are assessed under each benchmark's metrics and expert models. Notably, with the exception of T2ICompBench++, the prompt distributions of all other benchmarks do not overlap with our training data, indicating that performance on these benchmarks reflects out-of-distribution generalization capability.


\begin{itemize} 
\item \textbf{T2I-CompBench++}: Consists of four prompt categories—Attribute, Layout, Non-spatial Relationship, and Complex Composition. Each category is evaluated using an appropriate expert model or metric tailored to its specific requirements.
The Attribute category is measured by the Disentangled BLIP-VQA metric, which evaluates image-prompt semantic alignment via binary VQA responses from BLIP \citep{li2022blip}. The Layout category is measured by a UniDet-based metric that integrates depth estimation \citep{ranftl2021depth} and object detection \citep{zhou2022unidet} to assess 2D/3D spatial relationships and object counting. The Non-spatial Relationship category is measured using the CLIP-T metric, and the Complex Composition category is scored by a weighted combination of the three aforementioned metrics, referred to as the 3-in-1 metric.
A distinguishing characteristic of this benchmark is deliberate inclusion of unfamiliar or counterintuitive compositional prompts such as \textit{"a white banana"} which poses a greater challenge to current text-to-image generation models and provides a more rigorous evaluation of compositional generalization.

\item \textbf{DPGBench}: Comprises five prompt categories—Global, Entity, Attribute, Relation, and Other. It is distinguished by its use of notably longer prompts compared to other benchmarks. Evaluation is conducted uniformly across all categories using the mPLUG-large \citep{ye2023mplug} model as the adjudicator, which assesses generated images against a set of designated questions. Following the scoring protocol of DSG \citep{cho2023davidsonian}, the prompt score is computed as the mean of all question scores associated with a single prompt, and the overall DPG-Bench score is defined as the mean of all prompt scores.

\item \textbf{GenEval}: Evaluates compositional image generation across six categories—Single Object, Two Object, Counting, Colors, Position, and Attribute Binding. A Mask2Former \citep{cheng2022masked} model detects objects to verify presence, count, color (via CLIP \citep{radford2021clip}-based zero-shot classification), and relative spatial position depending on the category. Each image receives a binary correctness score, averaged per task and then across all tasks to yield the final score.
\end{itemize} 


\section{Additional Analysis}
\label{appendix:e-analyses}
To thoroughly evaluate OSPO, we conduct a series of analyses and ablation studies using the Janus-Pro-7B model. We begin by examining broader properties of OSPO, including prompt diversity, object mask quality, and image generation quality. We then evaluate its generalization ability to an alternative backbone model. Next, we analyze each stage of the OSPO framework. In the data construction stage, we examine the effect of prompt densification and the influence of the alignment-score threshold used for preference-pair filtering. In the training stage, we examine the effect of the key hyperparameter $\beta$, compare SimPO with DPO, and evaluate iterative training to verify OSPO's consistent self-boosting capability.
Beyond these components, we investigate the presence of confirmation bias—a commonly noted limitation in self-improving systems—by comparing the self-rewarding model against an external VQA-based reward model. Finally, we explore changes in the model's internal dynamics to provide further interpretability.



\begin{table}[t!]
\centering
{\fontsize{9}{11}\selectfont
\setlength{\tabcolsep}{0.8mm}
\begin{tabular}{ccccc|c}
\toprule
\textbf{Metric} &
\multicolumn{4}{c|}{\textbf{Category}} &
\textbf{Overall} \\
\cmidrule(lr){2-5}
 & Attribute & Layout & Non-Spatial & Complex & \\
\midrule
Self-BLEU     & 0.176 & 0.089 &  0.190 & 0.313 & 0.067 \\
\bottomrule
\end{tabular}
}
\caption{Self-BLEU across categories, measuring lexical diversity of generated prompts. Lower is better.}
\label{tab:self-bleu}
\end{table}

\label{appendix:e3-bleu}
\subsection{Prompt Diversity under Self-Generation}
To quantify prompt diversity, we report the lexical diversity of our generated prompts across categories by Self-BLEU \citep{zhu2018texygen} which is a metric to evaluate the diversity of the generated data. As shown in  Table~\ref{tab:self-bleu}, the overall Self-BLEU indicates low n-gram overlap across prompts, confirming that the generated prompts are not templated or repetitive. At the category level, the Layout category achieves the lowest Self-BLEU, reflecting high structural variety in spatial descriptions, while the Complex category shows the highest Self-BLEU, which is expected given that complex prompts inherently share more compositional patterns. Overall, these results demonstrate that our prompt generation produces lexically diverse training prompts across all categories.




\begin{table*}[t!]
\centering
\caption{Downstream benchmark performance of OSPO with self-generated masks and Grounded-SAM-2 masks. $\uparrow$ indicates higher is better, with bold highlighting the best score.}
{\fontsize{9}{11}\selectfont
\setlength{\tabcolsep}{2.7mm}
\begin{tabular}{l c cccc|c|c}  
\toprule
\multicolumn{1}{c}{\textbf{Model}} &
\textbf{Size} &
\multicolumn{4}{c|}{\textbf{T2I-CompBench++}$\uparrow$} &
\textbf{DPGBench$\uparrow$} &
\textbf{GenEval$\uparrow$} \\
\cmidrule(lr){3-6}
 &  & \textbf{Attribute} & \textbf{Layout} & \textbf{Non-Spatial} & \textbf{Complex} & \textbf{Overall} & \textbf{Overall} \\
\midrule
Janus-Pro                 & 7B & 0.65 & 0.41 & 0.31 & 0.36 & 81.87 & 0.796 \\
+ OSPO w/ self mask (default) & 7B & 0.756 & 0.447 & 0.316 & \textbf{0.415} & \textbf{85.61} & 0.830 \\
+ OSPO w/ GroundedSAM2 mask   & 7B & \textbf{0.769} & \textbf{0.449} & \textbf{0.317} & 0.407 & 85.45 & \textbf{0.831} \\
\bottomrule
\end{tabular}
}
\label{tab:mask-analysis}
\end{table*}

\subsection{mIoU of Object Mask against GT}
\label{appendix:e2-miou}
To evaluate whether our attention-derived object masks accurately localize the target objects, we measure mIoU against reference segmentation masks produced by Grounded-SAM-2 \citep{ren2024groundedsam} (GroundingDINO-Base + SAM2). We sample 500 training instances per category from iteration 1 and compute mIoU between OSPO masks and the corresponding Grounded-SAM-2 masks generated using the same set of object words. OSPO masks achieve an average mIoU of \textbf{0.5475} for preferred images and \textbf{0.5284} for non-preferred images, indicating that the attention-derived masks substantially overlap with detector-guided segmentations and capture meaningful object regions rather than arbitrarily covering background areas. Furthermore, as shown in Table~\ref{tab:mask-analysis}, the downstream performance gap between OSPO with self-generated masks and OSPO with Grounded-SAM-2 masks is negligible across all benchmarks, demonstrating that our self-supervised masking approach is a practical and effective alternative to external segmentation models.


\begin{table*}[t!]
\centering
\caption{Performance comparison of the Unitok-MLLM-8B on T2I-CompBench++, DPGBench, and GenEval. $\uparrow$ indicates higher is better, with bold highlighting the best score.}
{\fontsize{9}{11}\selectfont
\setlength{\tabcolsep}{4.2mm}
\begin{tabular}{l c cccc|c|c}  
\toprule
\multicolumn{1}{c}{\textbf{Model}} &
\textbf{Size} &
\multicolumn{4}{c|}{\textbf{T2I-CompBench++}$\uparrow$} &
\textbf{DPGBench$\uparrow$} &
\textbf{GenEval$\uparrow$} \\
\cmidrule(lr){3-6}
 &  & \textbf{Attribute} & \textbf{Layout} & \textbf{Non-Spatial} & \textbf{Complex} & \textbf{Overall} & \textbf{Overall} \\
\midrule
Unitok-MLLM               & 8B & 0.646 & \textbf{0.414} & 0.313 & 0.362 & 81.87 & 0.590 \\
\textbf{+ OSPO}    & 8B & \textbf{0.677} & \textbf{0.414} & \textbf{0.316} & \textbf{0.379} & \textbf{83.34} & \textbf{0.620} \\
\bottomrule
\end{tabular}
}
\label{tab:abl-unitok}
\end{table*}

\subsection{General T2I Generation Quality}
\label{appendix:e3-clip-lpips}
\begin{table}[t]
\centering
\caption{Comparison on MSCOCO-30K \citep{chen2015microsoft}. 
$\uparrow$ indicates higher is better, with bold highlighting the best score.}
{\fontsize{9}{11}\selectfont
\begin{tabular}{lcccc}
\toprule
\textbf{Model} & \textbf{Size} & \textbf{CLIP-T}$\uparrow$ & \textbf{CLIP-I}$\uparrow$ & \textbf{LPIPS}$\downarrow$ \\
\midrule
Janus-Pro  & 1B & 0.222 & 0.661 & 0.7137 \\
+ SILMM         & 1B & 0.227 & 0.667 & 0.7159 \\
+ \textbf{OSPO (ours)}   & 1B & \textbf{0.235} & \textbf{0.684} & \textbf{0.6996} \\
\midrule
Janus-Pro  & 7B & 0.235 & 0.704 & 0.7060 \\
+ SILMM       & 7B & 0.237 & \textbf{0.707} & 0.7023 \\
+ FocusDiff   & 7B & 0.232 & 0.672 & - \\
+ \textbf{OSPO (ours)} & 7B & \textbf{0.239} & \textbf{0.707} & \textbf{0.6959} \\
\bottomrule
\end{tabular}
} 
\label{tab:coco}
\end{table}

To assess the robustness of our framework under various input formats, we evaluate it on extensive compositional T2I benchmarks. Furthermore, we also measure CLIP-T, CLIP-I \citep{radford2021clip}, and LPIPS \citep{zhang2018lpips} score on the MSCOCO-30K \citep{chen2015microsoft} dataset to examine its general T2I generation capability. As reported in Table~\ref{tab:coco}, despite being designed for fine-grained semantic supervision, OSPO exhibits improvements in general T2I performance as well. This stands in contrast to FocusDiff \citep{pan2025focusdiff}, another fine-grained approach, which exhibits a performance decline on the same task.

\subsection{Effect of Backbone Model}
\label{appendix:e4-backbone}
To investigate the model-agnostic applicability of our framework, we conduct additional experiments using alternative MLLMs: UniTok-MLLM-8B \citep{ma2025unitok} with a different architecture. The model exhibits notably weaker image understanding and reasoning capabilities compared to Janus-Pro-7B, which are critical for our data generation pipeline. As shown in Table \ref{tab:abl-unitok}, applying OSPO consistently improves performance across different model architectures, demonstrating the framework’s generalizability and effectiveness.

\subsection{Effect of Prompt Densification}
\label{appendix:e5-densification}
\begin{figure}[t!]
    \centering
    \includegraphics[width=\linewidth]{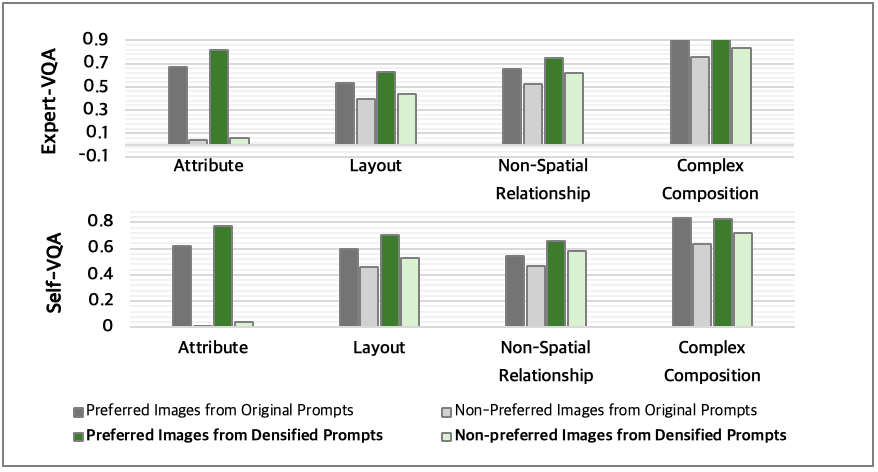}
    \caption{Mean alignment score computed over all alignment scores $S$ for preference pairs generated from original prompts (gray) and densified prompts (green), reported separately for each category. “Expert-VQA” reports scores estimated by the expert model, where as “Self-VQA” reports scores estimated by the model itself under the OSPO framework.}
\label{fig:abl-alignment-score}
\end{figure}

\begin{figure*}[t!]
    \centering
    \includegraphics[width=\linewidth]{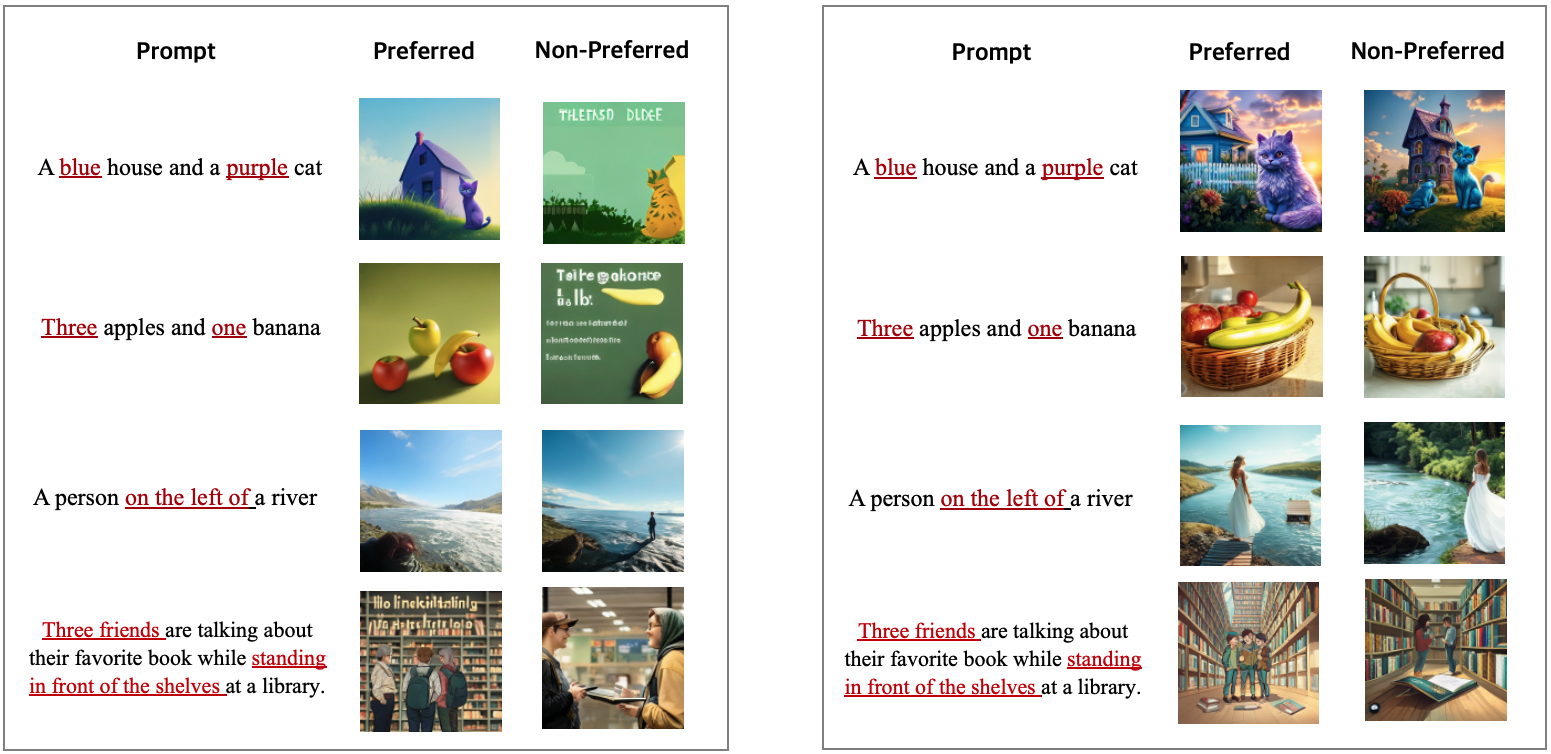}
    \caption{(Left) Training data examples whose images are generated from non-densified (original) prompts. (Right) Training data examples whose images are generated from densified prompts.}
\label{fig:abl-densification}
\end{figure*}

We have examined the effect of prompt densification on benchmark performance in Section~\ref{sec:analysis_dense} of the main paper. Here, we additionally evaluate the intrinsic image quality  by examining how faithfully the generated outputs capture the input text, comparing images synthesized from the original prompts with those synthesized from densified prompts. To measure fidelity at a fine-grained level, we use the same decompositional VQA protocol described in Section~\ref{sec:decompositional-vqa} of the main paper, querying a Vision-Language Model with atomic semantic questions. Using the two preference-pair datasets constructed in Section~\ref{sec:analysis_dense}—(1) the baseline set whose images are generated from the original prompts, and (2) our OSPO-constructed set generated from densified prompts.
We then evaluate their text–image alignment using the alignment score $S$ assigned by the expert VQA model InternVL3.5-14B \citep{wang2025internvl3}.
As shown in Figure~\ref{fig:abl-alignment-score}, prompt densification consistently improves the fidelity of preferred images across all categories, and even non-preferred images exhibit modest gains. These results indicate that our densification enhances the overall semantic consistency of generated images. Qualitative examples illustrating these visual differences are shown in Figure~\ref{fig:abl-densification}.


\begin{figure}[t!]
    \centering
    \includegraphics[width=\linewidth]{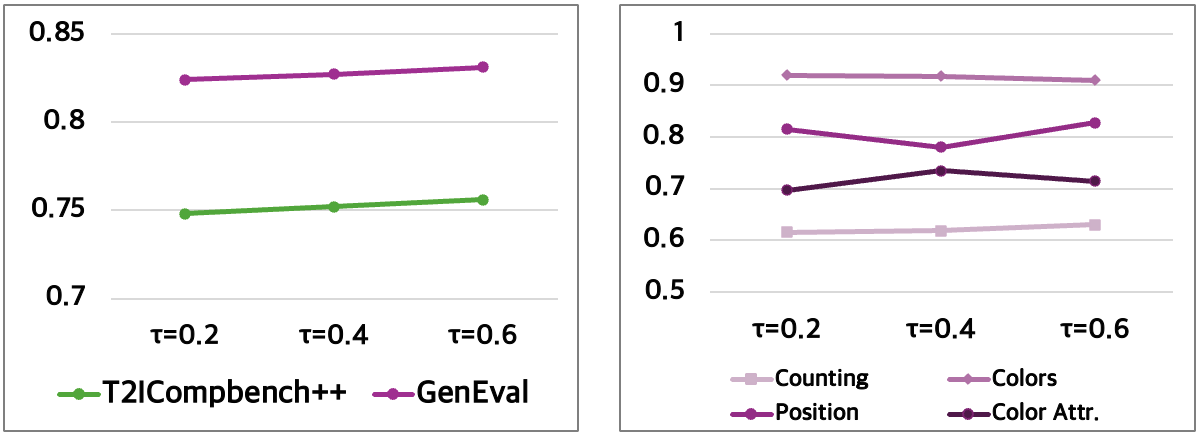}
    \caption{Effect of different filtering threshold ($\tau$) values. (Left) Performance comparison on T2I-CompBench++ (attribute) and GenEval (overall). (Right) Category-wise GenEval results. Higher is better for all benchmarks.}
\label{fig:abl-tau}
\end{figure}

\begin{figure}[t!]
    \centering
    \includegraphics[width=\linewidth]{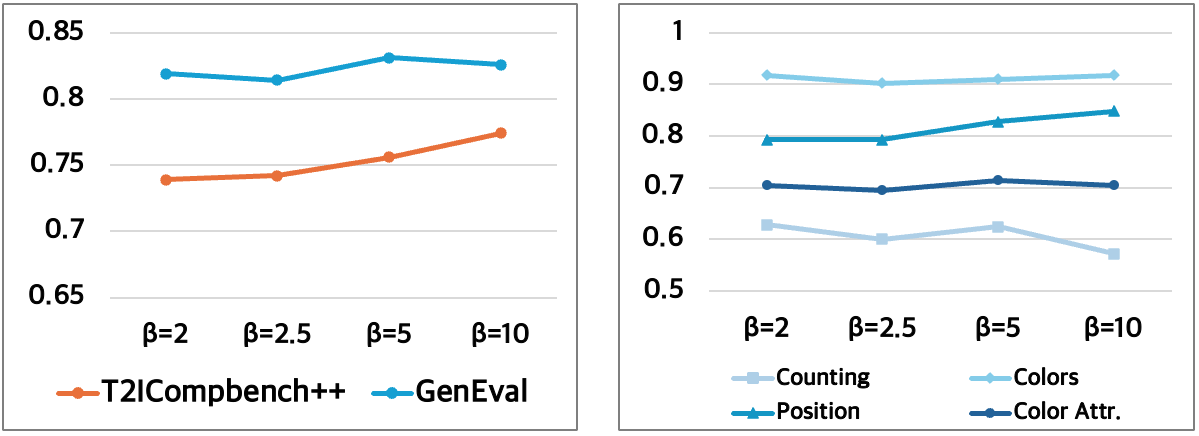}
    \caption{Effect of different $\beta$ values in SimPO. (Left) Performance comparison on T2I-CompBench++ (attribute) and GenEval (overall). (Right) Category-wise GenEval results. Higher is better for all benchmarks.}
\label{fig:abl-beta}
\end{figure}

\subsection{Effect of Filtering Threshold}
\label{appendix:e6-tau}
In the preference-pair construction stage, we filter samples by requiring the preferred image to have an alignment score $S$ above a threshold $\tau$. To evaluate the influence of this filtering criterion, we vary $\tau$ and compare the resulting training performance. Our default threshold is $\tau = 0.6$. As shown in Figure~\ref{fig:abl-tau}, increasing the threshold slightly improves benchmark performance. However, this must be considered alongside the fact that a higher threshold reduces the size of the training dataset.


\begin{table*}[t!]
\centering
\caption{Performance comparison of SimPO and DPO training within the OSPO framework on T2I-CompBench++, DPGBench, and GenEval. $\uparrow$ indicates higher is better, with bold highlighting the best score.}
{\fontsize{9}{11}\selectfont
\setlength{\tabcolsep}{3.2mm}
\begin{tabular}{l c cccc|c|c}  
\toprule
\multicolumn{1}{c}{\textbf{Model}} &
\textbf{Size} &
\multicolumn{4}{c|}{\textbf{T2I-CompBench++}$\uparrow$} &
\textbf{DPGBench$\uparrow$} &
\textbf{GenEval$\uparrow$} \\
\cmidrule(lr){3-6}
 &  & \textbf{Attribute} & \textbf{Layout} & \textbf{Non-Spatial} & \textbf{Complex} & \textbf{Overall} & \textbf{Overall} \\
\midrule
Janus-Pro                 & 7B & 0.418 & 0.292 & 0.311 & 0.387 & 81.87 & 0.796 \\
+ OSPO w/ SimPO (default)     & 7B & \textbf{0.756} & 0.447 & \textbf{0.316} & \textbf{0.415} & \textbf{85.61} & \textbf{0.830} \\
+ OSPO w/ DPO    & 7B & 0.700 & \textbf{0.448} & 0.315 & 0.400 & 84.80 & 0.808 \\
\bottomrule
\end{tabular}
}
\label{tab:abl-dpo}
\end{table*}


\begin{table*}[t!]
\centering
\caption{Performance improvement of Janus-Pro-7B on T2I-CompBench++, DPGBench, and GenEval with OSPO over three iterations. $\uparrow$ indicates higher is better, with bold highlighting the best score.}
{\fontsize{9}{11}\selectfont
\setlength{\tabcolsep}{4.2mm}
\begin{tabular}{l c cccc|c|c}  
\toprule
\multicolumn{1}{c}{\textbf{Model}} &
\textbf{Size} &
\multicolumn{4}{c|}{\textbf{T2I-CompBench++}$\uparrow$} &
\textbf{DPGBench$\uparrow$} &
\textbf{GenEval$\uparrow$} \\
\cmidrule(lr){3-6}
 &  & \textbf{Attribute} & \textbf{Layout} & \textbf{Non-Spatial} & \textbf{Complex} & \textbf{Overall} & \textbf{Overall} \\
\midrule
Janus-Pro                 & 7B & 0.418 & 0.292 & 0.311 & 0.387 & 81.87 & 0.796 \\
+ OSPO (Iter. 1)          & 7B & 0.756 & 0.447 & 0.316 & 0.415 & 85.61 & 0.830 \\
+ OSPO (Iter. 2)          & 7B & 0.756 & \textbf{0.455} & 0.317 & 0.412 & \textbf{86.00} & 0.831 \\
+ OSPO (Iter. 3)          & 7B & \textbf{0.760} & \textbf{0.455} & \textbf{0.318} & \textbf{0.418} & 85.84 & \textbf{0.835} \\
\bottomrule
\end{tabular}
}
\label{tab:abl-iter}
\end{table*}

\subsection{Effect of Training Strategy}
\label{appendix:e7-dpo}
Within the OSPO framework, we train our models using the SimPO ~\citep{meng2024simpo} algorithm, a simplified variant of DPO \citep{rafailov2023dpo} that removes the need for a reference model. In SimPO, the hyperparameter~$\beta$ scales the reward margin between preferred and non-preferred images, controlling the strength of preference supervision.
\paragraph{Hyperparameter $\beta$.}
In Preference Optimization, the hyperparameter $\beta$ is known to strongly influence optimization dynamics \citep{wu2024beta}. To understand its impact, we conduct an ablation study varying $\beta$ and analyze how different values affect training results. Figure~\ref{fig:abl-beta} shows that increasing $\beta$ improves performance in the Attribute category, but also raises the risk of reduced counting accuracy. Thus, $\beta$ must be carefully balanced.

\paragraph{Training Algorithm.}
Instead of SimPO, we train Janus-Pro-7B using the standard DPO algorithm \citep{rafailov2023dpo} while keeping our object-weighted token masking strategy unchanged. For DPO, we set $\beta=0.5$, a learning rate of 1e-5, a total batch size of 128, and 500 training steps; all other configurations follow the default OSPO training setup. As shown in Table~\ref{tab:abl-dpo}, DPO  also demonstrates noticeable improvements across benchmarks; however, SimPO consistently achieves larger performance gains. That's because it removes the reference-model constraint and directly optimizes the reward gap, enabling stronger object-level corrections and model updates that are essential for compositional text-image alignment.

\subsection{Effect of Iterative Training}
\label{appendix:e8-iterative-training}

As shown in Table~\ref{tab:abl-iter}, we present the progression of model performances across training iterations. The results indicate that the OSPO framework leads to steady and meaningful improvements in text-image alignment across a wide range of compositional categories, with a particularly notable improvement observed in the first iteration and continuing through subsequent iterations. 


\begin{table}[t!]
\centering
{\fontsize{9}{11}\selectfont
\setlength{\tabcolsep}{1.2mm}
\begin{tabular}{ccccc|c}
\toprule
\textbf{Metric} &
\multicolumn{4}{c|}{\textbf{Category}} &
\textbf{Overall} \\
\cmidrule(lr){2-5}
 & Attribute & Layout & Non-Spatial & Complex & \\
\midrule
Hit Rate     & 0.91 & 0.80 &  0.85 & 0.90 & 0.87 \\
\bottomrule
\end{tabular}
}
\caption{Hit rate across categories, measuring agreement between the self-rewarding model and the expert VQA model.}
\label{tab:abl-model-bias-vqa-score}
\end{table}
\subsection{Effect of Confirmation Bias}
\label{appendix:e9-bias}
In the OSPO framework, single MLLM serves both as a generator and a reward estimator, following the self-improving principle of using internal feedback to guide learning. In this section, we evaluate the validity of using the MLLM itself, specifically Janus-Pro-7B, as a reward estimator within the framework, by comparing its reward outputs with those of a relatively strong vision-language model, InternVL3.5-14B \citep{wang2025internvl3}. In the OSPO framework, the VQA-based alignment score $S$ serves as a reward output.

Table \ref{tab:abl-model-bias-vqa-score} reports the hit rate, defined as the proportion of matching binary answers (“yes” or “no”) between the self-model and the expert VQA model for the same candidate images. The hit rates remain high across categories (approximately 85\%), indicating that the self-model’s VQA outputs are largely consistent with those of the stronger model and therefore reasonably reliable. The largest discrepancy appears in the Layout category, which is expected since many MLLMs—including the self-model—tend to struggle with spatial reasoning.


\begin{table*}[t!]
\centering
\caption{Performance comparison of models trained with different VQA reward supervision on T2I-CompBench++, DPGBench, and GenEval. $\uparrow$ indicates higher is better, with bold highlighting the best score.}
{\fontsize{9}{11}\selectfont
\setlength{\tabcolsep}{2.85mm}
\begin{tabular}{l c cccc|c|c}  
\toprule
\multicolumn{1}{c}{\textbf{Model}} &
\textbf{Size} &
\multicolumn{4}{c|}{\textbf{T2I-CompBench++}$\uparrow$} &
\textbf{DPGBench$\uparrow$} &
\textbf{GenEval$\uparrow$} \\
\cmidrule(lr){3-6}
 &  & \textbf{Attribute} & \textbf{Layout} & \textbf{Non-Spatial} & \textbf{Complex} & \textbf{Overall} & \textbf{Overall} \\
\midrule
Janus-Pro                & 7B & 0.65 & 0.41 & 0.31 & 0.36 & 81.87 & 0.796 \\
+ OSPO w/ self-reward (default)     & 7B &\textbf{0.756} & \textbf{0.447} & \textbf{0.316} & \textbf{0.415} & 85.61 & \textbf{0.830} \\
+ OSPO w/ expert-reward   & 7B & \textbf{0.756} & 0.440 & \textbf{0.316} & 0.411 & \textbf{85.84} & 0.824 \\
\bottomrule
\end{tabular}
}
\label{tab:abl-model-bias}
\end{table*}

To assess whether differences in hit rate influence downstream training, we constructed a training dataset in which the VQA model used during filtering was replaced with the expert model \citep{wang2025internvl3}. As shown in Table~\ref{tab:abl-model-bias}, the models trained on datasets filtered by the self-model (OSPO) and by the expert model achieve nearly identical performance across all benchmarks. This suggests that using the model itself as the VQA-based reward estimator does not introduce meaningful confirmation bias during data filtering.

\subsection{Internal Model Behavior}
\label{appendix:e10-internal-analysis}

\begin{figure}[t!]
    \centering
    \includegraphics[width=\linewidth]{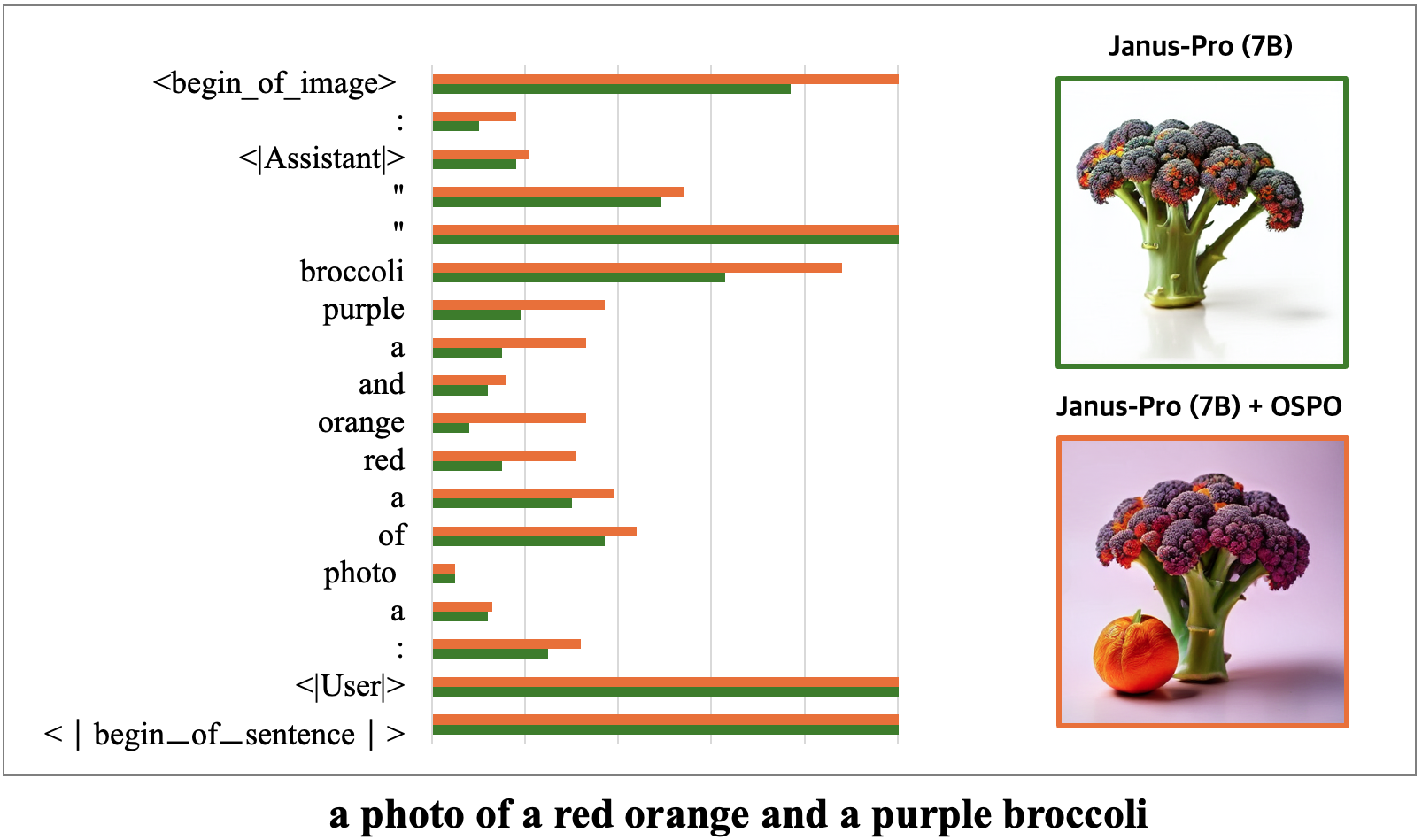}
    \caption{Comparison of globally averaged self-attention over text tokens during image-token generation for the baseline and OSPO-trained models. The baseline model fails to capture key semantic cues in the prompt, whereas the OSPO-trained model correctly reflects all intended semantics in the generated image.}
\label{fig:abl-attention}
\end{figure}

We analyze the model’s internal behavior by examining its self-attention distribution over input text tokens. As shown in Figure~\ref{fig:abl-attention}, OSPO increases attention on text tokens associated with key objects and attributes, indicating that the model becomes more grounded in the essential semantic elements of the prompt.

\begin{figure}[t!]
    \centering
    \includegraphics[width=\linewidth]{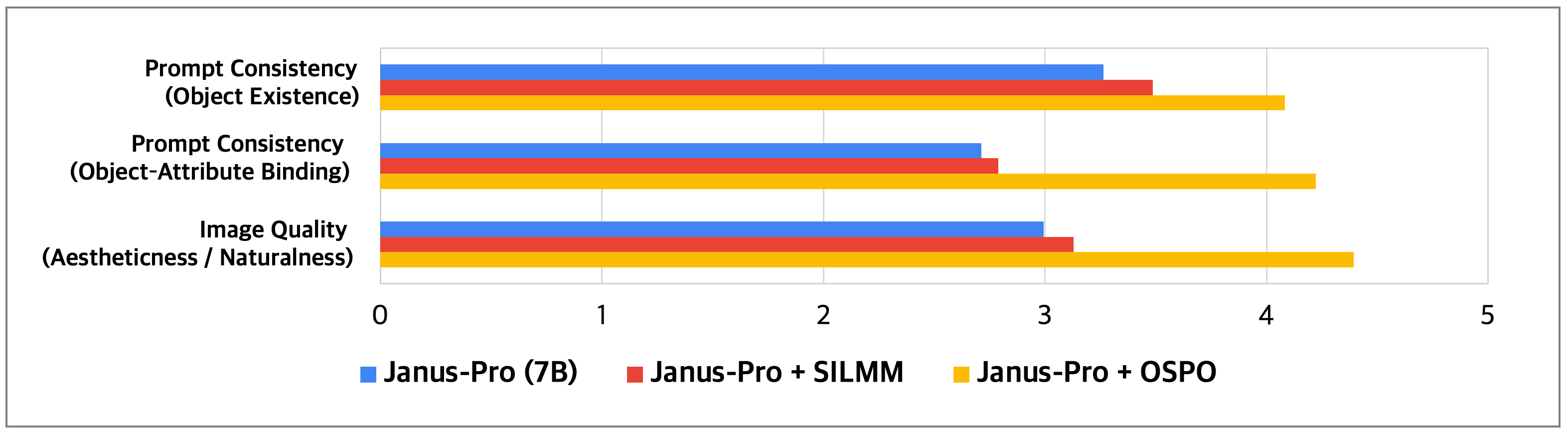}
    \caption{Human evaluation on 20 participants comparing images generated by the original Janus-Pro-7B, SILMM, and OSPO.}
\label{fig:human_eval}
\end{figure}

\section{Human Evaluation}
\label{appendix:f-human-evaluation}

\paragraph{Settings.}
We conducted a user study to validate our model's compositional generation quality under human perception. A total of 27 questions were presented to 20 participants, each consisting of one input text prompt and three generated images produced from the same prompt by different models. Participants rated each image on a 5-point scale (1: very poor, 5: very good) across three criteria: \textbf{Prompt Consistency (Object Existence)}, which evaluates whether all objects mentioned in the prompt are present in the generated image; \textbf{Prompt Consistency (Object–Attribute Binding)}, which assesses whether object attributes such as color, shape, texture, and spatial relations are correctly reflected; and \textbf{Image Quality (Aestheticness / Naturalness)}, which measures overall visual quality including the absence of artifacts and naturalness of the output. The three compared models are the original Janus-Pro, SILMM, and OSPO.

\paragraph{Results.}
 As shown in Figure~\ref{fig:human_eval}, OSPO outperforms both the original model and SILMM across all three criteria. The most pronounced improvements are observed in the two prompt consistency criteria, demonstrating that OSPO substantially enhances the model's ability to generate images that faithfully reflect the semantics of the input prompt. OSPO also achieves the highest image quality score, indicating that improved compositional alignment does not come at the cost of visual naturalness.

\section{Qualitative Evaluation}
\label{appendix:g-qualitative}

\begin{figure*}[t!]
    \centering
    \includegraphics[width=\linewidth]{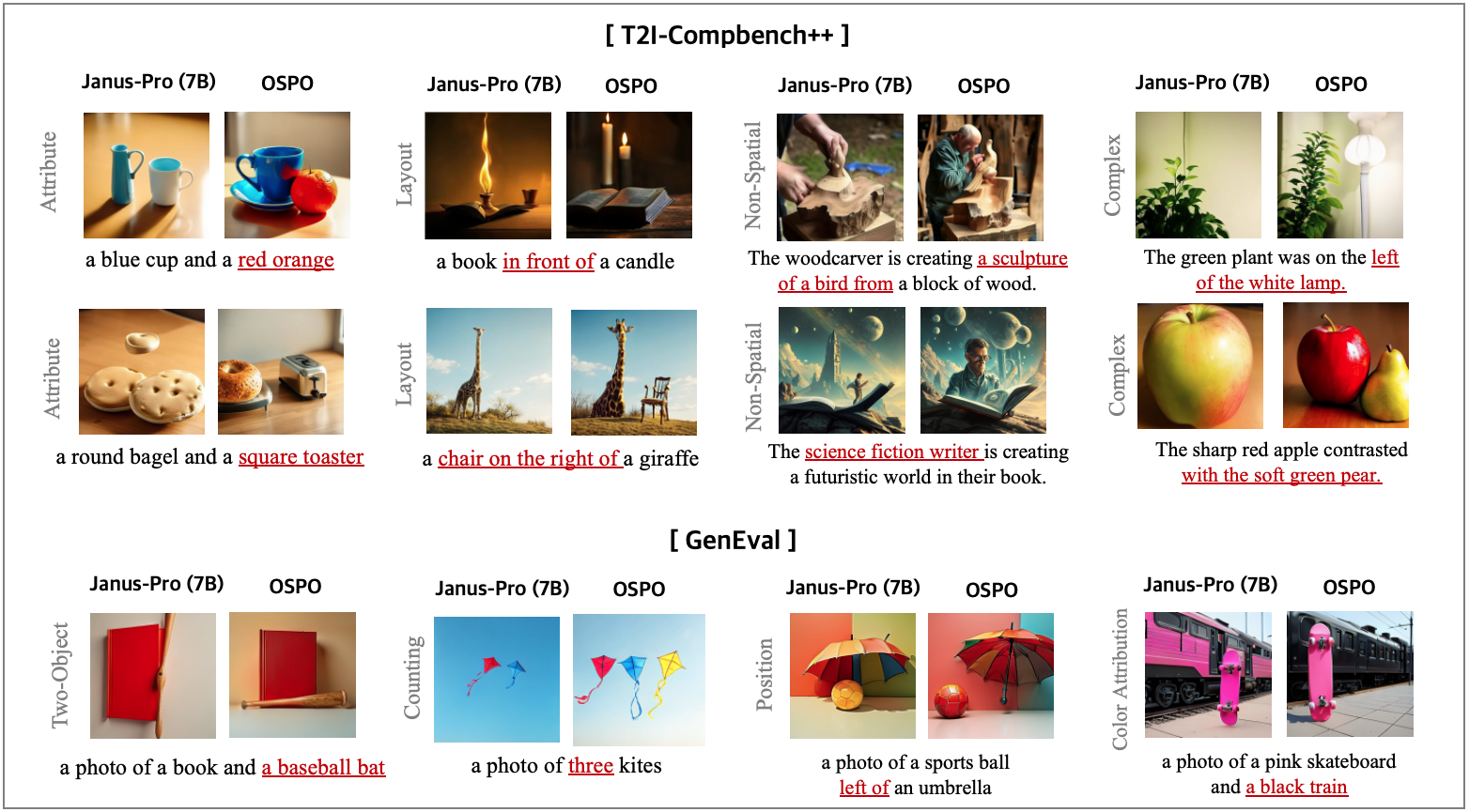}
    \caption{Additional Qualitative results of Janus-Pro-7B \citep{chen2025januspro} and Janus-Pro-7B + OSPO on T2I-CompBench++ \citep{huang2025t2icompbench} and GenEval \citep{ghosh2023geneval}.}
\label{fig:janus_pro_7b_t2i_geneval}
\end{figure*}

\begin{figure*}[t!]
    \centering
    \includegraphics[width=\linewidth]{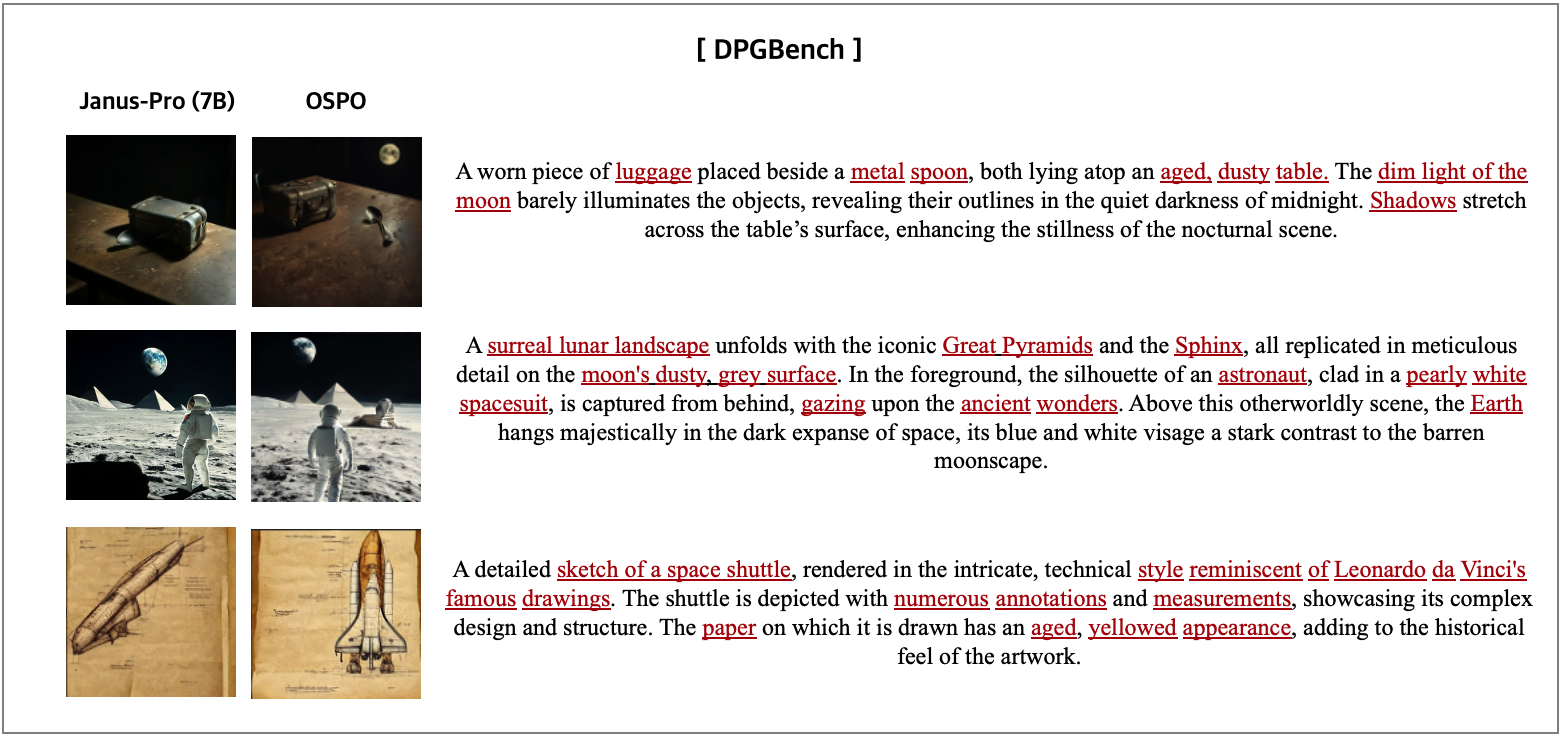}
    \caption{Additional Qualitative results of Janus-Pro-7B \citep{chen2025januspro} and Janus-Pro-7B + OSPO on DPGBench \citep{hu2024dpgbench}.}
\label{fig:janus_pro_7b_dpg}
\end{figure*}

\begin{figure*}[t!]
    \centering
    \includegraphics[width=\linewidth]{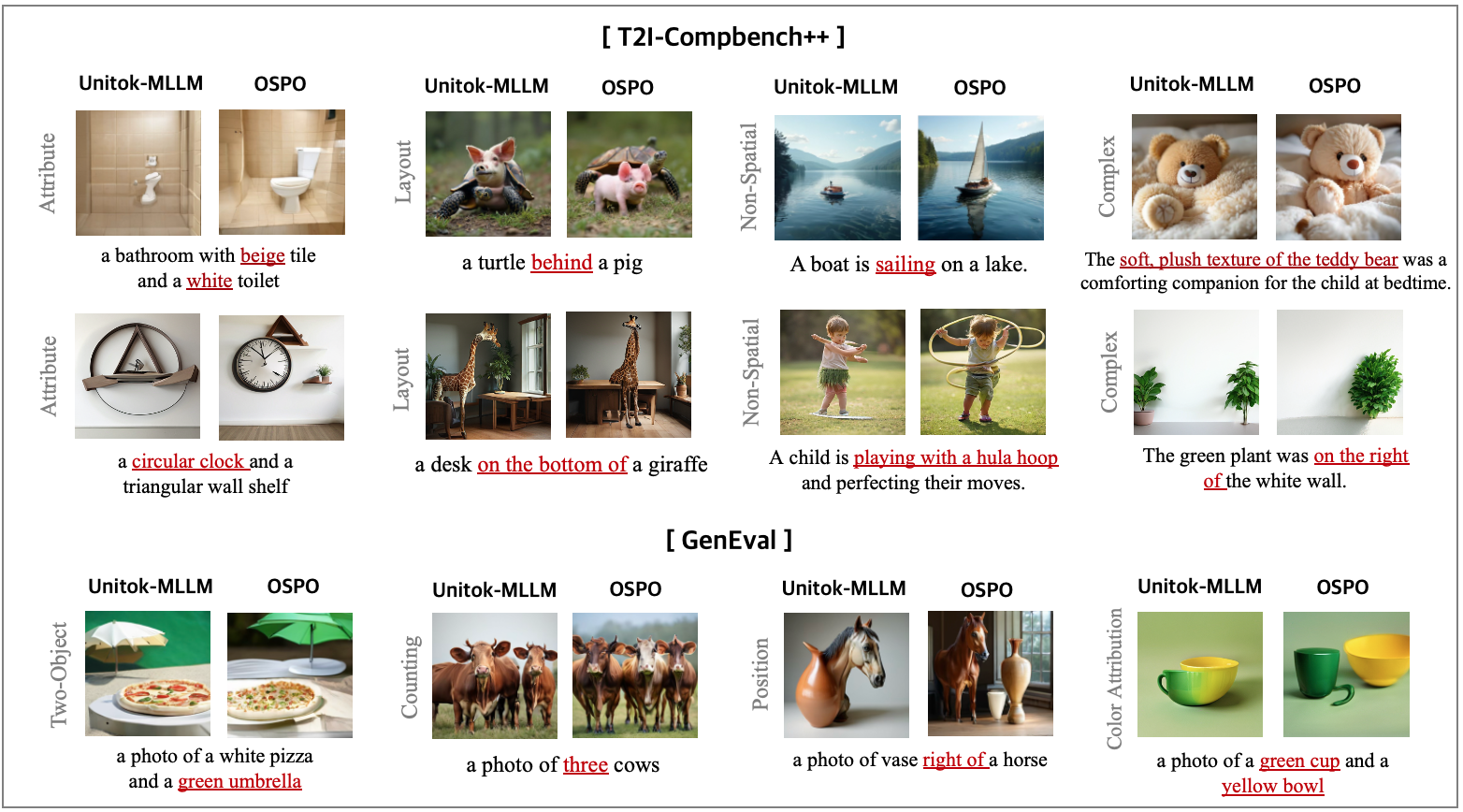}
    \caption{Additional Qualitative results of UniTok-MLLM-8B \citep{ma2025unitok} and UniTok-MLLM-8B + OSPO on T2I-CompBench++ \citep{huang2025t2icompbench} and GenEval \citep{ghosh2023geneval}.}
\label{fig:unitok_t2i_geneval}
\end{figure*}

\begin{figure*}[t!]
    \centering
    \includegraphics[width=\linewidth]{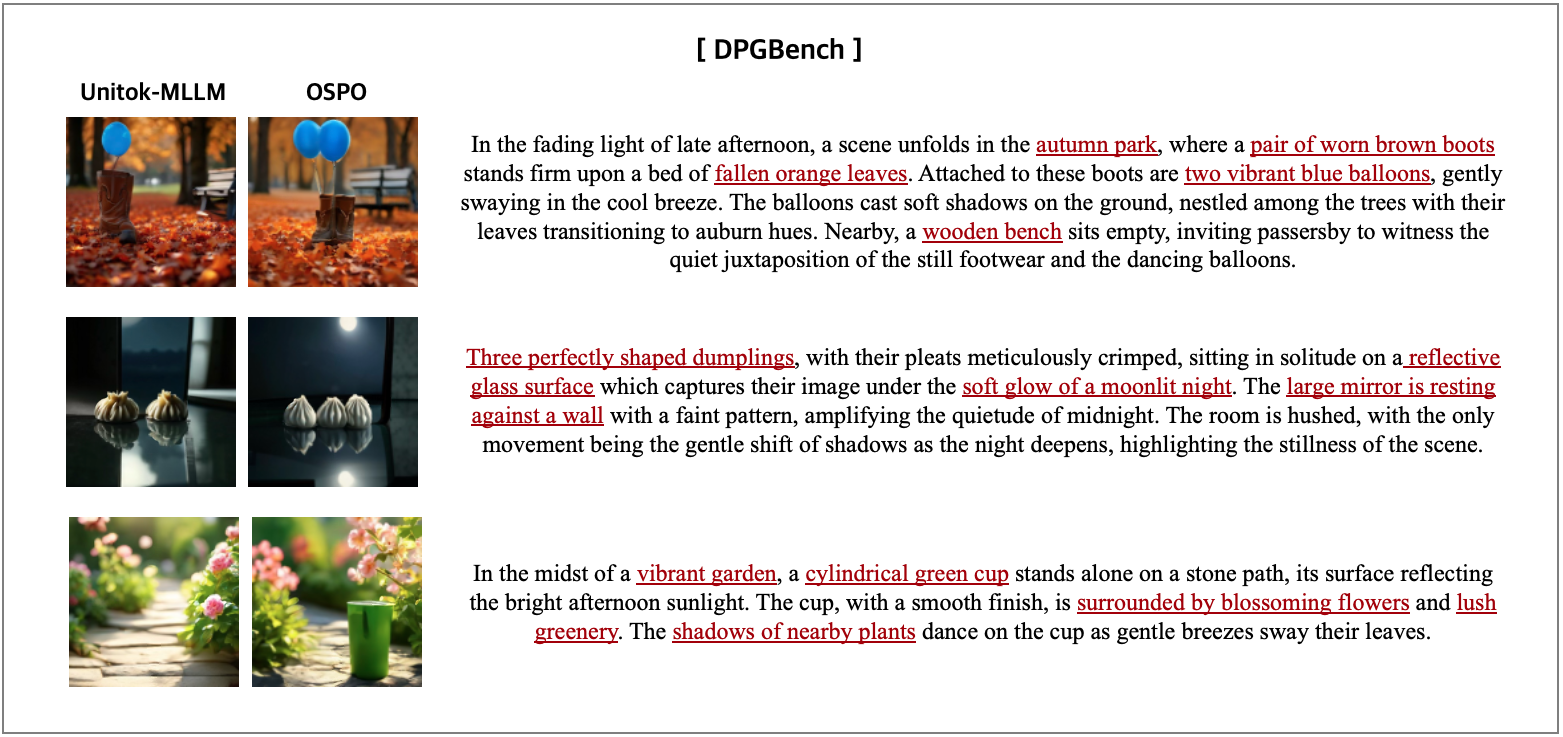}
    \caption{Additional Qualitative results of UniTok-MLLM-8B \citep{ma2025unitok} and UniTok-MLLM-8B + OSPO on DPGBench \citep{hu2024dpgbench}.}
\label{fig:unitok_dpg}
\end{figure*}

We present additional visual examples to qualitatively and intuitively compare each baseline (Janus-Pro-1B, Janus-Pro-7B, and UniTok-MLLM-8B) with the corresponding results after training by OSPO. 
Figures \ref{fig:janus_pro_7b_t2i_geneval} and \ref{fig:janus_pro_7b_dpg} present comparisons between Janus-Pro-7B and Janus-Pro-7B + OSPO on T2I-CompBench++, GenEval, and DPGBench. 
Figures \ref{fig:unitok_t2i_geneval} and \ref{fig:unitok_dpg} show comparisons between UniTok-MLLM-8B and UniTok-MLLM-8B + OSPO on the T2I-CompBench++, GenEval, and DPGBench as well. 

As shown in the results, the baseline models Janus-Pro-1B, Janus-Pro-7B, and UniTok-MLLM-8B often fail to follow the prompt accurately, exhibiting object omissions, incorrect attribute bindings, and overall misalignment between the text prompt and the generated image. In contrast, training with OSPO framework leads to improved alignment, with models better preserving object presence and correctly binding attributes. These results highlight the effectiveness of OSPO in mitigating object-level hallucinations.

\section{Prompt Data Examples}
\label{appendix:h-prompt-examples}

In this section, we present representative examples of data generated throughout the OSPO framework. Table~\ref{tab:base_prompt} summarizes the category-specific base prompt formats and examples. Since the Non-spatial and Complex categories do not follow a specific prompt format; therefore, formats for these are omitted in the table. Notably, our baseline SILMM \citep{qu2024silmm} constructs its training data using the same base prompt format derived from T2I-Compbench++, ensuring a fair and directly comparable evaluation. 
Table~\ref{tab:prompt-perturb} provides examples of perturbed prompts across the three perturbation types, and Table~\ref{tab:prompt-densify} provides the corresponding densified prompts derived from the Attribute example. 


\section{Prompt Templates}
\label{appendix:h-prompt-templates}

We leverage the In-Context Learning (ICL) \citep{dong2022icl} capabilities of MLLMs throughout the OSPO framework. Specifically, we first generate original prompts corresponding to four categories—Attribute, Layout, Non-Spatial, and Complex Composition. Using these original prompts, we construct different kinds of perturbed prompts, which are then used to produce dense prompts. During the Preference Pair Construction stage, we also generate VQA questions to perform Decompositional VQA, again leveraging the ICL capabilities of the MLLMs.

\section{Future Works}
Self-improving frameworks can be extended to more advanced capabilities, including self-reasoning and inference-time scaling. Recent studies \citep{Duan2025GoT-R1, Pan2025JanusProR1, Zhang2025ReasonGenR1} show that incorporating multimodal reasoning processes—such as self-refinement during image generation—can further improve T2I performance. As future work, we plan to integrate such reasoning mechanisms into OSPO to enhance its generation ability. Another promising direction is to extend OSPO beyond image generation by incorporating image understanding tasks, thereby moving toward a more unified and efficient self-improving framework for MLLMs.

\newpage
\begin{table*}[t!]
\centering
\caption{Generated base prompt formats and example prompts across Attribute, Layout, Non-spatial, and Complex categories.}
\setlength{\tabcolsep}{0.5mm}
\begin{tabular}{p{3cm} p{7cm} p{7cm}}
\toprule
\textbf{Category} & \textbf{Prompt Format} & \textbf{Base Prompt} \\
\midrule
\textbf{Attribute} & a \{adj\} \{obj\} & a green dress \\
 & a \{adj1\} \{obj1\} and a \{adj2\} \{obj2\} & a fabric pillow and a ceramic cup \\
\midrule
\textbf{Layout} & a \{noun1\} \{spatial\} a \{noun2\} & a phone on the right side of a bag \\
& \{quantity\} \{obj(s)\} & two trees \\
& \{quantity\} \{obj1(s)\} and
\{quantity\} \{obj2(s)\} & four cookies and two umbrellas \\
\midrule
\textbf{Non-Spatial} & - & A person is holding a book while sitting on a bench in a park. \\
\midrule
\textbf{Complex} & - & The deep crimson, velvet curtains framed the large, cream-colored window and the luxurious, gold-trimmed mirrors reflected the warm, ambient lighting of the elegant room. \\
\bottomrule
\end{tabular}
\label{tab:base_prompt}
\end{table*}

\begin{table*}[t!]
\centering
\caption{Examples of perturbed prompts for the Replace, Swap, and Drop strategies. Underlined text marks the portion that differs semantically from the original prompt.}
\setlength{\tabcolsep}{0.5mm}
\begin{tabular}{%
    p{3cm}
    p{6cm}
    @{\hspace{6mm}}   
    p{7.5cm}
}
\toprule
\textbf{Category} & \textbf{Base Prompt} & \textbf{Perturbed Prompt} \\
\midrule
\textbf{Attribute} & a blue house and a purple cat & (Replace) a \underline{green} house and a \underline{yellow} cat \\
& & (Swap) a \underline{purple} house and a \underline{blue} cat \\
& & (Drop) a \underline{blue} house \\
\midrule
\textbf{Layout} & a book above a dog & (Replace)  a book \underline{under} a dog \\
& & (Swap) a \underline{dog} above a \underline{book} \\
& & (Drop) a dog \underline{and} a book \\
\midrule
\textbf{Non-Spatial} & A group of friends are playing soccer in a park on a sunny day. & (Replace)  A group of friends are \underline{riding bicycles} in a park on a sunny day. \\
& & (Drop) A group of friends \underline{are playing} in a park on a sunny day. \\
\midrule
\textbf{Complex} & The vibrant, multicolored beads shimmered below the clear, polished glass vase. & (Replace)  The \underline{dark, monochrome} beads shimmered below the transparent, glossy vase. \\
& & (Swap) The clear, polished glass \underline{vase} shimmered \underline{below} the vibrant, multicolored \underline{beads}. \\
& & (Drop) The plain, red, yellow beads are floating.\\
\bottomrule
\end{tabular}
\label{tab:prompt-perturb}
\end{table*}

\begin{table*}[t!]
\centering
\caption{Examples of densified prompts generated from the base and perturbed prompts. Underlined text highlights the differing semantic units.}
\setlength{\tabcolsep}{0.5mm}
\begin{tabular}{p{7cm} p{1cm} p{8.8cm}}
\toprule
\textbf{Original Prompt} & & \textbf{Densified Prompt} \\
\midrule
(Base) a blue house and a purple cat & & A quaint, \underline{blue-painted house} with a charming porch and a cheerful \underline{purple cat} lounging on the wooden steps, watching a soft spring breeze rustle the nearby wildflowers. \\
\midrule
(Perturbed) a purple house and a blue cat & & A quaint, \underline{purple-painted house} with a charming porch and a cheerful \underline{blue cat} lounging on the wooden steps, watching a soft spring breeze rustle the nearby wildflowers. \\
\bottomrule
\end{tabular}
\label{tab:prompt-densify}
\end{table*}

\clearpage
\begin{tcolorbox}[breakable, colback=black!5!white,colframe=gray!75!black,title=\textbf{Base Prompt Generation Template}]

\textbf{System Prompt:} You are an assistant dedicated to generating natural prompts that contain subjects and objects by using color words such as red, orange, yellow, green, blue, navy, purple. Do not repeat any previously used colors in subsequent prompts.\\

\textbf{User:} Generate a prompt that contains subjects and objects by using color words. Except for the color generated before, do not use the same color.\\
\textbf{Assistant:} Prompt: a blue bench and a green car.; Concepts and relations: blue bench, green car; Relation: (bench, blue), (car, green)\\

\textbf{User:} Generate a prompt that contains subjects and objects by using color words. Except for the color generated before, do not use the same color.\\
\textbf{Assistant:} Prompt: a red elephant and a white grass.; Concepts and relations: red elephant, white grass; Relation: (elephant, red), (grass, white)\\

\textbf{User:} Generate a prompt that contains subjects and objects by using color words. Except for the color generated before, do not use the same color.\\
\textbf{Assistant:} Prompt: a yellow pendant and an orange apple.; Concepts and relations: yellow pendant, orange apple; Relation: (pendant, yellow), (apple, orange)\\
\tcblower
\textbf{User:} Generate a prompt that contains subjects and objects by using color words. Except for the color generated before, do not use the same color.\\
\textbf{Assistant:}
\end{tcolorbox}

\begin{tcolorbox}[breakable, colback=black!5!white,colframe=gray!75!black,title=\textbf{Prompt Perturbation Template}]

\textbf{System Prompt:} You are a helpful assistant that generates object-level contrastive prompts by following four steps.\\
Step 1. Identify all objects in the prompt. \\
Step 2. Identify all object-attribute bindings in the prompt.\\
Step 3. Apply one of the following transformations to the given prompt: (1) `replace' an attribute or object with another plausible one, (2) `swap' two attributes or objects, or (3) `drop' an attribute or object.\\
Step 4. Generate contrastive prompts based on the transformations. \\
Always ensure the output has opposing or conflicting semantics compared to the input. Do not just reorder or repeat the input.\\

\textbf{\textbf{User:}} a blue bench and a red car\\
Generate object-level contrastive prompts by applying the following transformation: `replace'. Let's think step by step.\\
\textbf{Assistant:} Step 1. Objects: bench, car\\
Step 2. Attribute Bindings: [`blue', `bench'], [`red', `car']\\
Step 3. Transformation: `replace'; Transformed Attributes: [`white', `bench'], [`green', `car']\\
Step 4. Contrastive Prompt: a white bench and a green car\\

\textbf{User:} a rubber tire and a wooden desk\\
Generate object-level contrastive prompts by applying the following transformation: `swap'. Let's think step by step.\\
\textbf{Assistant:} Step 1. Objects: tire, desk\\
Step 2. Attribute Bindings: [`rubber', `tire'], [`wooden', `desk']\\
Step 3. Transformation: `swap'; Transformed Attributes: [`wooden', `tire'], [`rubber', `desk']\\
Step 4. Contrastive Prompt: a wooden tire and a rubber desk\\

\textbf{User:} a oblong sweet potato and teardrop mango\\
Generate object-level contrastive prompts by applying the following transformation: `drop'. Avoid changing the original attribute bindings of the objects that are not dropped. Let's think step by step.\\
\textbf{Assistant:} Step 1. Objects: sweet potato, mango\\
Step 2. Attributes Bindings: [`oblong', `sweet potato'], [`teardrop', `mango']\\
Step 3. Transformation: `drop'; Transformed Attributes: [`teardrop', `mango']\\
Step 4. Contrastive Prompt: a teardrop mango

\tcblower

\textbf{User:} \verb|{input_prompt}|\\
Generate object-level contrastive prompts by applying the following transformation: \verb|{perturbation_type}|. Let's think step by step.\\
\textbf{Assistant:}\\
\end{tcolorbox}

\begin{tcolorbox}[breakable, colback=black!5!white, colframe=gray!75!black, title=\textbf{Prompt Densification Template}]

\textbf{System Prompt:} You are an expert prompt engineer for text-to-image models. Your job is to take short and vague prompts and expand them into detailed, descriptive, and unambiguous prompts suitable for high-quality image generation.\\
Focus on using full sentences and include visual attributes such as objects, colors, texture, shape.\\
Avoid abstract or subjective words and instead use concrete and visual language.\\
Do not invent unrelated concepts; only expand and clarify the given prompt.\\
Follow these steps:\\
Step 1. Extract all objects and their visual attributes from Prompt 1.\\
Step 2. For Prompt 1, write a long, rich description that includes all identified objects and attributes from Step 1.\\
Step 3. Extract all objects and their visual attributes from Prompt 2.\\
Step 4. For Prompt 2, write a long, rich description that includes all identified objects and attributes from Step 3.\\
Ensure both outputs share a similar global context or scene.\\

\textbf{User:} Prompt 1: A large watermelon\\
Prompt 2: A small watermelon\\
Generate dense, detailed prompts. Ensure both outputs share a similar global context or scene but have different object-attribute bindings. Let's think step by step.\\
\textbf{Assistant:}\\
Step 1. Prompt 1 Object Bindings: watermelon-['large']\\
Step 2. Prompt 1 Dense: A large, ripe watermelon with deep green rinds and faint striping rests heavily in a handwoven wicker basket placed on the grass under dappled sunlight beside a weathered garden shed.\\
Step 3. Prompt 2 Object Bindings: watermelon-['small']\\
Step 4. Prompt 2 Dense: A small, round watermelon with bright green skin and subtle mottling sits neatly in a handwoven wicker basket placed on the grass under dappled sunlight beside a weathered garden shed.\\

\textbf{User:} Prompt 1: A peach tree and a square computer keyboard\\
Prompt 2: An orange tree and a square keyboard\\
Generate dense, detailed prompts. Ensure both outputs share a similar global context or scene but have different object-attribute bindings. Let's think step by step.\\
\textbf{Assistant:}\\
Step 1. Prompt 1 Object Bindings: tree-['peach']; keyboard-['square', 'computer']\\
Step 2. Prompt 1 Dense: A peach tree in full bloom stands beside a sleek, square-shaped computer keyboard resting on a polished wooden desk, with soft morning sunlight streaming through the windows of a quiet home office.\\
Step 3. Prompt 2 Object Bindings: tree-['orange']; keyboard-['square']\\
Step 4. Prompt 2 Dense: An orange tree with lush green leaves and dangling fruit stands beside a sleek, square-shaped computer keyboard resting on a polished wooden desk, with soft morning sunlight streaming through the windows of a quiet home office.\\

\textbf{User:} Prompt 1: A long in length elephant and a teal owl\\
Prompt 2: A short elephant and a blue owl\\
Generate dense, detailed prompts. Let's think step by step.\\
\textbf{Assistant:}\\
Step 1. Prompt 1 Object Bindings: elephant-['long in length']; owl-['teal']\\
Step 2. Prompt 1 Dense: A long-bodied elephant with weathered gray skin trudging slowly through a misty jungle path, accompanied by a serene teal owl perched on a mossy branch overhead, its feathers glimmering in the filtered morning light.\\
Step 3. Prompt 2 Object Bindings: elephant-['short']; owl-['blue']\\
Step 4. Prompt 2 Dense: A short, stocky elephant with wrinkled gray skin navigating a misty jungle path, accompanied by a curious blue owl perched on a mossy branch overhead, its feathers catching flecks of golden morning light.

\tcblower

\textbf{User:} Prompt 1: \verb|{base_input_prompt}|\\
Prompt 2: \verb|{negative_input_prompt}|\\
Generate dense, detailed prompts. Ensure both outputs share a similar global context or scene but have different object-attribute bindings. Let's think step by step.\\
\textbf{Assistant:}
\end{tcolorbox}

\begin{tcolorbox}[colback=black!5!white,colframe=gray!75!black,title=\textbf{Decompositional Question Generation Template}]
\textbf{System Prompt:} You are an assistant dedicated to transforming a sentence into several questions. You should first divide it into simple concepts and relations, and then provide the corresponding questions. Avoid using pronouns, such as he, she, it, and they.\\

\textbf{User:} A white harp and a rust soup.\\
\textbf{Assistant:} Concepts and relations: a white harp, a rust soup; Questions: Is there a white harp? Is there a rust soup?\\

\textbf{User:} A quarter circle lily and a hexagon mirror.\\
\textbf{Assistant:} Concepts and relations: a quarter circle lily, a hexagon mirror; Questions: Is there a quarter-circle lily? Is there a hexagon mirror? \\

\textbf{User:} Shiny mop and metal key holder.\\
\textbf{Assistant:} Concepts and relations: a shiny mop, a metal key holder; Questions: Is there a shiny mop? Is there a metal key holder?

\tcblower

\textbf{User:} \verb|{input_prompt}| \\
\textbf{Assistant:}
\end{tcolorbox}

\begin{figure*}[t!]
    \centering
    \includegraphics[width=\linewidth]{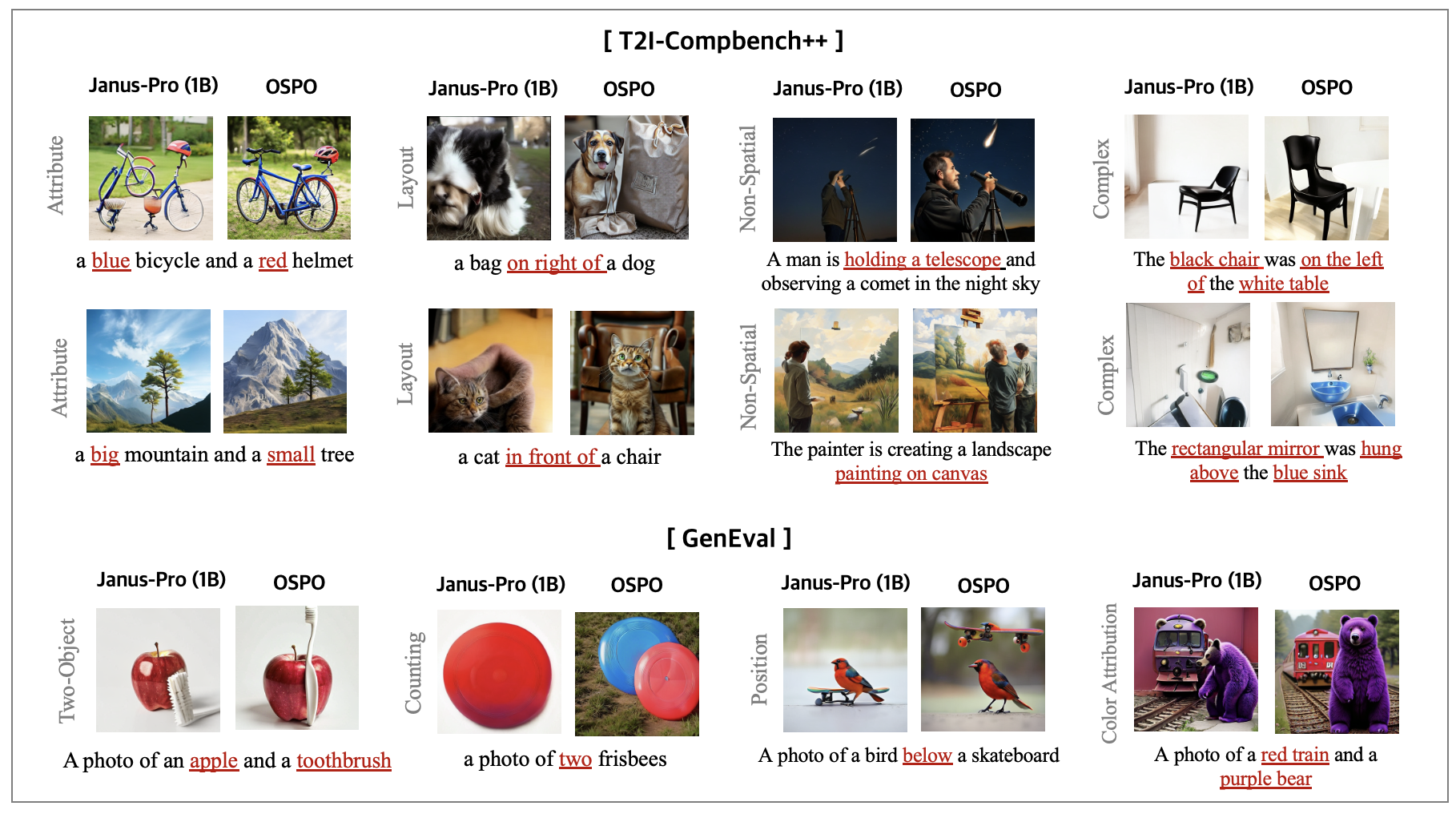}
    \caption{Additional Qualitative results of Janus-Pro-1B \citep{chen2025januspro} and Janus-Pro-1B + OSPO on T2I-CompBench++ \citep{huang2025t2icompbench} and GenEval \citep{ghosh2023geneval}.}
\label{fig:janus_pro_1b_t2i_geneval}
\end{figure*}

\begin{figure*}[t!]
    \centering
    \includegraphics[width=\linewidth]{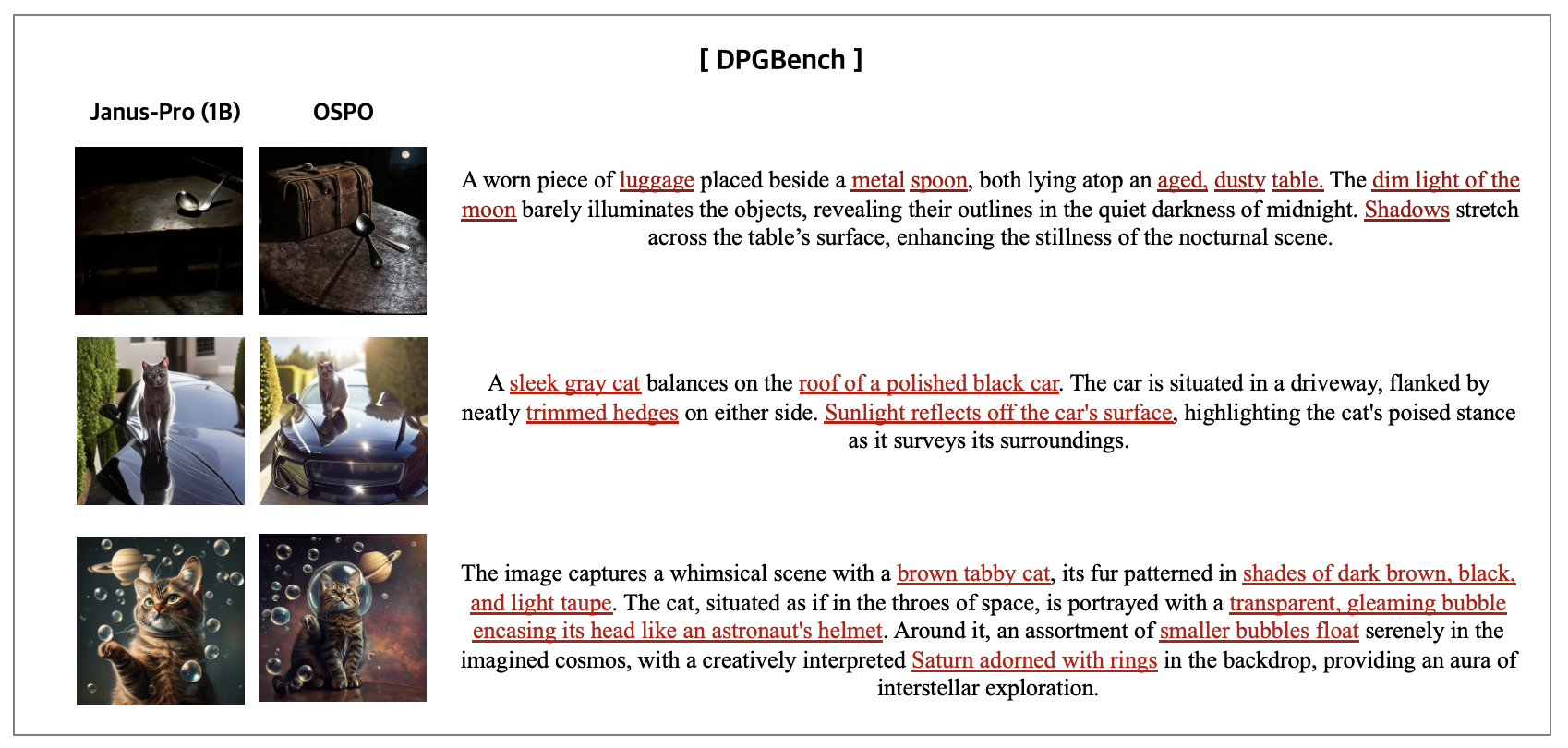}
    \caption{Additional Qualitative results of Janus-Pro-1B \citep{chen2025januspro} and Janus-Pro-1B + OSPO on DPGBench \citep{hu2024dpgbench}.}
\label{fig:janus_pro_1b_dpg}
\end{figure*}

\clearpage
\twocolumn
{
    \small
    \bibliographystyle{ieeenat_fullname}
    \bibliography{supp_refs}

@String(ICLR = {Int. Conf. Learn. Represent.})

@String(ICLR  = {ICLR})

@article{team2024chameleon,
  title={Chameleon: Mixed-modal early-fusion foundation models},
  author={Team, Chameleon},
  journal={arXiv preprint arXiv:2405.09818},
  year={2024}
}

@article{sun2023emu1,
  title={Emu: Generative pretraining in multimodality},
  author={Sun, Quan and Yu, Qiying and Cui, Yufeng and Zhang, Fan and Zhang, Xiaosong and Wang, Yueze and Gao, Hongcheng and Liu, Jingjing and Huang, Tiejun and Wang, Xinlong},
  journal={arXiv preprint arXiv:2307.05222},
  year={2023}
}

@inproceedings{sun2024emu2,
  title={Generative multimodal models are in-context learners},
  author={Sun, Quan and Cui, Yufeng and Zhang, Xiaosong and Zhang, Fan and Yu, Qiying and Wang, Yueze and Rao, Yongming and Liu, Jingjing and Huang, Tiejun and Wang, Xinlong},
  booktitle={Proceedings of the IEEE/CVF Conference on Computer Vision and Pattern Recognition},
  pages={14398--14409},
  year={2024}
}

@article{wang2024emu3,
  title={Emu3: Next-token prediction is all you need},
  author={Wang, Xinlong and Zhang, Xiaosong and Luo, Zhengxiong and Sun, Quan and Cui, Yufeng and Wang, Jinsheng and Zhang, Fan and Wang, Yueze and Li, Zhen and Yu, Qiying and others},
  journal={arXiv preprint arXiv:2409.18869},
  year={2024}
}

@article{ge2023seedllama,
  title={Making llama see and draw with seed tokenizer},
  author={Ge, Yuying and Zhao, Sijie and Zeng, Ziyun and Ge, Yixiao and Li, Chen and Wang, Xintao and Shan, Ying},
  journal={arXiv preprint arXiv:2310.01218},
  year={2023}
}

@article{wu2024vilau,
  title={Vila-u: a unified foundation model integrating visual understanding and generation},
  author={Wu, Yecheng and Zhang, Zhuoyang and Chen, Junyu and Tang, Haotian and Li, Dacheng and Fang, Yunhao and Zhu, Ligeng and Xie, Enze and Yin, Hongxu and Yi, Li and others},
  journal={arXiv preprint arXiv:2409.04429},
  year={2024}
}

@article{xie2024show1,
  title={Show-o: One single transformer to unify multimodal understanding and generation},
  author={Xie, Jinheng and Mao, Weijia and Bai, Zechen and Zhang, David Junhao and Wang, Weihao and Lin, Kevin Qinghong and Gu, Yuchao and Chen, Zhijie and Yang, Zhenheng and Shou, Mike Zheng},
  journal={arXiv preprint arXiv:2408.12528},
  year={2024}
}

@article{xie2025show2,
  title={Show-o2: Improved Native Unified Multimodal Models},
  author={Xie, Jinheng and Yang, Zhenheng and Shou, Mike Zheng},
  journal={arXiv preprint arXiv:2506.15564},
  year={2025}
}

@article{liu2023llava,
  title={Visual instruction tuning},
  author={Liu, Haotian and Li, Chunyuan and Wu, Qingyang and Lee, Yong Jae},
  journal={Advances in neural information processing systems},
  volume={36},
  pages={34892--34916},
  year={2023}
}

@article{ye2023mplug,
  title={mplug-owl: Modularization empowers large language models with multimodality},
  author={Ye, Qinghao and Xu, Haiyang and Xu, Guohai and Ye, Jiabo and Yan, Ming and Zhou, Yiyang and Wang, Junyang and Hu, Anwen and Shi, Pengcheng and Shi, Yaya and others},
  journal={arXiv preprint arXiv:2304.14178},
  year={2023}
}

@article{dai2023instructblip,
  title={Instructblip: Towards general-purpose vision-language models with instruction tuning},
  author={Dai, Wenliang and Li, Junnan and Li, Dongxu and Tiong, Anthony and Zhao, Junqi and Wang, Weisheng and Li, Boyang and Fung, Pascale N and Hoi, Steven},
  journal={Advances in neural information processing systems},
  volume={36},
  pages={49250--49267},
  year={2023}
}

@inproceedings{chen2024internvl,
  title={Internvl: Scaling up vision foundation models and aligning for generic visual-linguistic tasks},
  author={Chen, Zhe and Wu, Jiannan and Wang, Wenhai and Su, Weijie and Chen, Guo and Xing, Sen and Zhong, Muyan and Zhang, Qinglong and Zhu, Xizhou and Lu, Lewei and others},
  booktitle={Proceedings of the IEEE/CVF conference on computer vision and pattern recognition},
  pages={24185--24198},
  year={2024}
}

@inproceedings{wu2025janus,
  title={Janus: Decoupling visual encoding for unified multimodal understanding and generation},
  author={Wu, Chengyue and Chen, Xiaokang and Wu, Zhiyu and Ma, Yiyang and Liu, Xingchao and Pan, Zizheng and Liu, Wen and Xie, Zhenda and Yu, Xingkai and Ruan, Chong and others},
  booktitle={Proceedings of the Computer Vision and Pattern Recognition Conference},
  pages={12966--12977},
  year={2025}
}

@article{chen2025januspro,
  title={Janus-pro: Unified multimodal understanding and generation with data and model scaling},
  author={Chen, Xiaokang and Wu, Zhiyu and Liu, Xingchao and Pan, Zizheng and Liu, Wen and Xie, Zhenda and Yu, Xingkai and Ruan, Chong},
  journal={arXiv preprint arXiv:2501.17811},
  year={2025}
}

@article{ma2025unitok,
  title={Unitok: A unified tokenizer for visual generation and understanding},
  author={Ma, Chuofan and Jiang, Yi and Wu, Junfeng and Yang, Jihan and Yu, Xin and Yuan, Zehuan and Peng, Bingyue and Qi, Xiaojuan},
  journal={arXiv preprint arXiv:2502.20321},
  year={2025}
}

@article{schulman2017proximal,
  title={Proximal policy optimization algorithms},
  author={Schulman, John and Wolski, Filip and Dhariwal, Prafulla and Radford, Alec and Klimov, Oleg},
  journal={arXiv preprint arXiv:1707.06347},
  year={2017}
}

@article{rafailov2023dpo,
  title={Direct preference optimization: Your language model is secretly a reward model},
  author={Rafailov, Rafael and Sharma, Archit and Mitchell, Eric and Manning, Christopher D and Ermon, Stefano and Finn, Chelsea},
  journal={Advances in Neural Information Processing Systems},
  volume={36},
  pages={53728--53741},
  year={2023}
}

@article{meng2024simpo,
  title={Simpo: Simple preference optimization with a reference-free reward},
  author={Meng, Yu and Xia, Mengzhou and Chen, Danqi},
  journal={Advances in Neural Information Processing Systems},
  volume={37},
  pages={124198--124235},
  year={2024}
}

@article{shao2024deepseekmath,
  title={Deepseekmath: Pushing the limits of mathematical reasoning in open language models},
  author={Shao, Zhihong and Wang, Peiyi and Zhu, Qihao and Xu, Runxin and Song, Junxiao and Bi, Xiao and Zhang, Haowei and Zhang, Mingchuan and Li, YK and Wu, Yang and others},
  journal={arXiv preprint arXiv:2402.03300},
  year={2024}
}

@article{qu2024silmm,
  title={SILMM: Self-Improving Large Multimodal Models for Compositional Text-to-Image Generation},
  author={Qu, Leigang and Li, Haochuan and Wang, Wenjie and Liu, Xiang and Li, Juncheng and Nie, Liqiang and Chua, Tat-Seng},
  journal={arXiv preprint arXiv:2412.05818},
  year={2024}
}

@article{jiang2025t2ir1,
  title={T2i-r1: Reinforcing image generation with collaborative semantic-level and token-level cot},
  author={Jiang, Dongzhi and Guo, Ziyu and Zhang, Renrui and Zong, Zhuofan and Li, Hao and Zhuo, Le and Yan, Shilin and Heng, Pheng-Ann and Li, Hongsheng},
  journal={arXiv preprint arXiv:2505.00703},
  year={2025}
}

@article{mao2025unirl,
  title={UniRL: Self-Improving Unified Multimodal Models via Supervised and Reinforcement Learning},
  author={Mao, Weijia and Yang, Zhenheng and Shou, Mike Zheng},
  journal={arXiv preprint arXiv:2505.23380},
  year={2025}
}

@inproceedings{Hong2025SUDERSU,
  title={SUDER: Self-Improving Unified Large Multimodal Models for Understanding and Generation with Dual Self-Rewards},
  author={Jixiang Hong and Yiran Zhang and Guanzhong Wang and Yi Liu and Ji-Rong Wen and Rui Yan},
  year={2025},
  url={https://api.semanticscholar.org/CorpusID:279250731}
}

@article{pan2025focusdiff,
  title={FocusDiff: Advancing Fine-Grained Text-Image Alignment for Autoregressive Visual Generation through RL},
  author={Pan, Kaihang and Bu, Wendong and Wu, Yuruo and Wu, Yang and Shen, Kai and Li, Yunfei and Zhao, Hang and Li, Juncheng and Tang, Siliang and Zhuang, Yueting},
  journal={arXiv preprint arXiv:2506.05501},
  year={2025}
}

@article{van2017neural,
  title={Neural discrete representation learning},
  author={Van Den Oord, Aaron and Vinyals, Oriol and others},
  journal={Advances in neural information processing systems},
  volume={30},
  year={2017}
}

@inproceedings{esser2021taming,
  title={Taming transformers for high-resolution image synthesis},
  author={Esser, Patrick and Rombach, Robin and Ommer, Bjorn},
  booktitle={Proceedings of the IEEE/CVF conference on computer vision and pattern recognition},
  pages={12873--12883},
  year={2021}
}

@article{huang2025t2icompbench,
  title={T2I-CompBench++: An Enhanced and Comprehensive Benchmark for Compositional Text-to-Image Generation},
  author={Huang, Kaiyi and Duan, Chengqi and Sun, Kaiyue and Xie, Enze and Li, Zhenguo and Liu, Xihui},
  journal={IEEE Transactions on Pattern Analysis and Machine Intelligence},
  year={2025},
  publisher={IEEE}
}

@inproceedings{hu2023tifa,
  title={Tifa: Accurate and interpretable text-to-image faithfulness evaluation with question answering},
  author={Hu, Yushi and Liu, Benlin and Kasai, Jungo and Wang, Yizhong and Ostendorf, Mari and Krishna, Ranjay and Smith, Noah A},
  booktitle={Proceedings of the IEEE/CVF International Conference on Computer Vision},
  pages={20406--20417},
  year={2023}
}

@article{dong2022icl,
  title={A survey on in-context learning},
  author={Dong, Qingxiu and Li, Lei and Dai, Damai and Zheng, Ce and Ma, Jingyuan and Li, Rui and Xia, Heming and Xu, Jingjing and Wu, Zhiyong and Liu, Tianyu and others},
  journal={arXiv preprint arXiv:2301.00234},
  year={2022}
}

@article{hu2022lora,
  title={Lora: Low-rank adaptation of large language models.},
  author={Hu, Edward J and Shen, Yelong and Wallis, Phillip and Allen-Zhu, Zeyuan and Li, Yuanzhi and Wang, Shean and Wang, Lu and Chen, Weizhu and others},
  journal={ICLR},
  volume={1},
  number={2},
  pages={3},
  year={2022}
}

@inproceedings{thrush2022winoground,
  title={Winoground: Probing vision and language models for visio-linguistic compositionality},
  author={Thrush, Tristan and Jiang, Ryan and Bartolo, Max and Singh, Amanpreet and Williams, Adina and Kiela, Douwe and Ross, Candace},
  booktitle={Proceedings of the IEEE/CVF Conference on Computer Vision and Pattern Recognition},
  pages={5238--5248},
  year={2022}
}

@article{hsieh2023sugarcrepe,
  title={Sugarcrepe: Fixing hackable benchmarks for vision-language compositionality},
  author={Hsieh, Cheng-Yu and Zhang, Jieyu and Ma, Zixian and Kembhavi, Aniruddha and Krishna, Ranjay},
  journal={Advances in neural information processing systems},
  volume={36},
  pages={31096--31116},
  year={2023}
}

@article{yang2023idea2img,
  title={Idea2img: Iterative self-refinement with gpt-4v (ision) for automatic image design and generation},
  author={Yang, Zhengyuan and Wang, Jianfeng and Li, Linjie and Lin, Kevin and Lin, Chung-Ching and Liu, Zicheng and Wang, Lijuan},
  journal={arXiv preprint arXiv:2310.08541},
  year={2023}
}

@article{yang2025self,
  title={Self-Rewarding Large Vision-Language Models for Optimizing Prompts in Text-to-Image Generation},
  author={Yang, Hongji and Zhou, Yucheng and Han, Wencheng and Shen, Jianbing},
  journal={arXiv preprint arXiv:2505.16763},
  year={2025}
}

@inproceedings{wang2024promptcharm,
  title={Promptcharm: Text-to-image generation through multi-modal prompting and refinement},
  author={Wang, Zhijie and Huang, Yuheng and Song, Da and Ma, Lei and Zhang, Tianyi},
  booktitle={Proceedings of the 2024 CHI Conference on Human Factors in Computing Systems},
  pages={1--21},
  year={2024}
}

@inproceedings{li2024promptist,
  title={PROMPTIST: Automated prompt optimization for text-to-image synthesis},
  author={Li, WeiJie and Wang, Jin and Zhang, Xuejie},
  booktitle={CCF international conference on natural language processing and Chinese computing},
  pages={295--306},
  year={2024},
  organization={Springer}
}

@article{wang2024mdpo,
  title={mdpo: Conditional preference optimization for multimodal large language models},
  author={Wang, Fei and Zhou, Wenxuan and Huang, James Y and Xu, Nan and Zhang, Sheng and Poon, Hoifung and Chen, Muhao},
  journal={arXiv preprint arXiv:2406.11839},
  year={2024}
}

@article{hu2024dpgbench,
  title={Ella: Equip diffusion models with llm for enhanced semantic alignment},
  author={Hu, Xiwei and Wang, Rui and Fang, Yixiao and Fu, Bin and Cheng, Pei and Yu, Gang},
  journal={arXiv preprint arXiv:2403.05135},
  year={2024}
}

@article{cho2023davidsonian,
  title={Davidsonian scene graph: Improving reliability in fine-grained evaluation for text-to-image generation},
  author={Cho, Jaemin and Hu, Yushi and Garg, Roopal and Anderson, Peter and Krishna, Ranjay and Baldridge, Jason and Bansal, Mohit and Pont-Tuset, Jordi and Wang, Su},
  journal={arXiv preprint arXiv:2310.18235},
  year={2023}
}

@article{ghosh2023geneval,
  title={Geneval: An object-focused framework for evaluating text-to-image alignment},
  author={Ghosh, Dhruba and Hajishirzi, Hannaneh and Schmidt, Ludwig},
  journal={Advances in Neural Information Processing Systems},
  volume={36},
  pages={52132--52152},
  year={2023}
}

@article{wang2025internvl3,
  title={Internvl3. 5: Advancing open-source multimodal models in versatility, reasoning, and efficiency},
  author={Wang, Weiyun and Gao, Zhangwei and Gu, Lixin and Pu, Hengjun and Cui, Long and Wei, Xingguang and Liu, Zhaoyang and Jing, Linglin and Ye, Shenglong and Shao, Jie and others},
  journal={arXiv preprint arXiv:2508.18265},
  year={2025}
}

@inproceedings{radford2021clip,
  title={Learning transferable visual models from natural language supervision},
  author={Radford, Alec and Kim, Jong Wook and Hallacy, Chris and Ramesh, Aditya and Goh, Gabriel and Agarwal, Sandhini and Sastry, Girish and Askell, Amanda and Mishkin, Pamela and Clark, Jack and others},
  booktitle={International conference on machine learning},
  pages={8748--8763},
  year={2021},
  organization={PmLR}
}

@article{hessel2021clipscore,
  title={Clipscore: A reference-free evaluation metric for image captioning},
  author={Hessel, Jack and Holtzman, Ari and Forbes, Maxwell and Bras, Ronan Le and Choi, Yejin},
  journal={arXiv preprint arXiv:2104.08718},
  year={2021}
}

@article{chen2015microsoft,
  title={Microsoft coco captions: Data collection and evaluation server},
  author={Chen, Xinlei and Fang, Hao and Lin, Tsung-Yi and Vedantam, Ramakrishna and Gupta, Saurabh and Doll{\'a}r, Piotr and Zitnick, C Lawrence},
  journal={arXiv preprint arXiv:1504.00325},
  year={2015}
}

@article{otsu1975threshold,
  title={A threshold selection method from gray-level histograms},
  author={Otsu, Nobuyuki and others},
  journal={Automatica},
  volume={11},
  number={285-296},
  pages={23--27},
  year={1975}
}

@inproceedings{kang2025your,
  title={Your large vision-language model only needs a few attention heads for visual grounding},
  author={Kang, Seil and Kim, Jinyeong and Kim, Junhyeok and Hwang, Seong Jae},
  booktitle={Proceedings of the Computer Vision and Pattern Recognition Conference},
  pages={9339--9350},
  year={2025}
}

@inproceedings{binyamin2025make,
  title={Make it count: Text-to-image generation with an accurate number of objects},
  author={Binyamin, Lital and Tewel, Yoad and Segev, Hilit and Hirsch, Eran and Rassin, Royi and Chechik, Gal},
  booktitle={Proceedings of the Computer Vision and Pattern Recognition Conference},
  pages={13242--13251},
  year={2025}
}

@inproceedings{Pan2025JanusProR1,
  author    = {Kaihang Pan and Yang Wu and Wendong Bu and Kai Shen and Juncheng Li and Yingting Wang and Yunfei Li and Siliang Tang and Jun Xiao and Fei Wu and Hang Zhao and Yueting Zhuang},
  title     = {Janus-Pro-R1: Advancing Collaborative Visual Comprehension and Generation via Reinforcement Learning},
  booktitle = {Proceedings of the 39th Annual Conference on Neural Information Processing Systems (NeurIPS 2025)},
  year      = {2025},
  note      = {Poster},
}

@article{Duan2025GoT-R1,
  title        = {GoT-R1: Unleashing Reasoning Capability of MLLM for Visual Generation with Reinforcement Learning},
  author       = {Chengqi Duan and Rongyao Fang and Yuqing Wang and Kun Wang and Linjiang Huang and Xingyu Zeng and Hongsheng Li and Xihui Liu},
  journal      = {arXiv preprint arXiv:2505.17022},
  year         = {2025},
}

@article{Zhang2025ReasonGenR1,
  title        = {ReasonGen-R1: CoT for Autoregressive Image Generation Models through SFT and RL},
  author       = {Yu Zhang and Yunqi Li and Yifan Yang and Rui Wang and Yuqing Yang and Dai Qi and Jianmin Bao and Dongdong Chen and Chong Luo and Lili Qiu},
  journal      = {arXiv preprint arXiv:2505.24875},
  year         = {2025},
}

@article{wu2024beta,
  title={{$\beta$}-DPO: Direct Preference Optimization with Dynamic {$\beta$}},
  author={Wu, Junkang and Xie, Yuexiang and Yang, Zhengyi and Wu, Jiancan and Gao, Jinyang and Ding, Bolin and Wang, Xiang and He, Xiangnan},
  journal={Advances in Neural Information Processing Systems},
  volume={37},
  pages={129944--129966},
  year={2024}
}

@inproceedings{li2022blip,
  title={Blip: Bootstrapping language-image pre-training for unified vision-language understanding and generation},
  author={Li, Junnan and Li, Dongxu and Xiong, Caiming and Hoi, Steven},
  booktitle={International conference on machine learning},
  pages={12888--12900},
  year={2022},
  organization={PMLR}
}

@inproceedings{zhou2022unidet,
  title={Simple multi-dataset detection},
  author={Zhou, Xingyi and Koltun, Vladlen and Kr{\"a}henb{\"u}hl, Philipp},
  booktitle={Proceedings of the IEEE/CVF conference on computer vision and pattern recognition},
  pages={7571--7580},
  year={2022}
}

@inproceedings{ranftl2021depth,
  title={Vision transformers for dense prediction},
  author={Ranftl, Ren{\'e} and Bochkovskiy, Alexey and Koltun, Vladlen},
  booktitle={Proceedings of the IEEE/CVF international conference on computer vision},
  pages={12179--12188},
  year={2021}
}

@inproceedings{cheng2022masked,
  title={Masked-attention mask transformer for universal image segmentation},
  author={Cheng, Bowen and Misra, Ishan and Schwing, Alexander G and Kirillov, Alexander and Girdhar, Rohit},
  booktitle={Proceedings of the IEEE/CVF conference on computer vision and pattern recognition},
  pages={1290--1299},
  year={2022}
}

@article{ren2024groundedsam,
  title={Grounded sam: Assembling open-world models for diverse visual tasks},
  author={Ren, Tianhe and Liu, Shilong and Zeng, Ailing and Lin, Jing and Li, Kunchang and Cao, He and Chen, Jiayu and Huang, Xinyu and Chen, Yukang and Yan, Feng and others},
  journal={arXiv preprint arXiv:2401.14159},
  year={2024}
}

@inproceedings{zhu2018texygen,
  title={Texygen: A benchmarking platform for text generation models},
  author={Zhu, Yaoming and Lu, Sidi and Zheng, Lei and Guo, Jiaxian and Zhang, Weinan and Wang, Jun and Yu, Yong},
  booktitle={The 41st international ACM SIGIR conference on research \& development in information retrieval},
  pages={1097--1100},
  year={2018}
}

@inproceedings{zhang2018lpips,
  title={The unreasonable effectiveness of deep features as a perceptual metric},
  author={Zhang, Richard and Isola, Phillip and Efros, Alexei A and Shechtman, Eli and Wang, Oliver},
  booktitle={Proceedings of the IEEE conference on computer vision and pattern recognition},
  pages={586--595},
  year={2018}
}

}

\end{document}